\documentclass[Afour,sageh,times]{sagej}

\usepackage{graphicx}
\usepackage{float}
\usepackage{amsmath}
\usepackage{amssymb}

\let\vec\boldvec

\newcommand{\trsp}{\mathsf{T}}

\usepackage{epstopdf}

\usepackage{subfigure}
\usepackage{algorithm}
\usepackage{algpseudocode}
\usepackage{enumitem}

\usepackage{moreverb,url}

\usepackage{booktabs}       
\usepackage[toc,page]{appendix}
\DeclareMathOperator*{\argmax}{arg\,max}
\DeclareMathOperator*{\argmin}{arg\,min}
\usepackage{threeparttable, tablefootnote}

\usepackage{caption}
\usepackage{varwidth}
\DeclareCaptionFormat{varwidth}{%
	\begin{varwidth}{\linewidth}#1#2#3\end{varwidth}%
}
\usepackage[colorlinks,citecolor=red]{hyperref}

\newcommand\BibTeX{{\rmfamily B\kern-.05em \textsc{i\kern-.025em b}\kern-.08em
		T\kern-.1667em\lower.7ex\hbox{E}\kern-.125emX}}

\def\volumeyear{2016}

\begin{document}
	
\runninghead{Huang et al.}
	
\title{Kernelized Movement Primitives}
	
\author{Yanlong Huang\affilnum{1}, Leonel Rozo\affilnum{1}, Jo\~{a}o Silv\'{e}rio\affilnum{1} and Darwin G. Caldwell\affilnum{1}}
	
\affiliation{\affilnum{1} Department of Advanced Robotics, Istituto Italiano di Tecnologia.}
	
\corrauth{Yanlong Huang,
	Department of Advanced Robotics,
	Istituto Italiano di Tecnologia,
	Via Morego 30,
	16163 Genova, Italy.}
	
\email{yanlong.huang@iit.it}
	
\begin{abstract}
Imitation learning has been studied widely as a convenient way to transfer human skills to robots. This learning approach is aimed at extracting relevant motion patterns from human demonstrations and subsequently applying these patterns to different situations. 
Despite many advancements have been achieved, the solutions for coping with unpredicted situations (e.g., obstacles and external perturbations) and high-dimensional inputs are still largely open.
In this paper, we propose a novel \emph{kernelized movement primitive} (KMP), which allows the robot to adapt the learned motor skills and fulfill a variety of additional constraints arising over the course of a task. Specifically, KMP is capable of learning trajectories associated with high-dimensional inputs due to the kernel treatment, which in turn renders a model with fewer open parameters in contrast to methods that rely on basis functions. Moreover, we extend
our approach by 
exploiting local trajectory representations in different coordinate systems that describe the task at hand,
endowing KMP with reliable extrapolation capabilities in broader domains. We apply KMP to the learning of time-driven trajectories as a special case, where a compact parametric representation describing a trajectory and its first-order derivative is utilized. In order to verify the effectiveness of our method,  several examples of trajectory modulations and extrapolations associated with time inputs, as well as trajectory adaptations with high-dimensional inputs are provided.
\end{abstract}
	
	
\maketitle

\section{Introduction}
In a myriad of robotic systems, trajectory generation plays a very important role since trajectories govern the robot actions at both joint and task space levels. One popular trajectory generation approach for robots is imitation learning \citep{Ijspeert,Calinon2007}, where the trajectory of interest is learned from human demonstrations. Typically, the learned trajectories can be successfully reproduced or generalized by the robot under conditions that are similar to those in which the demonstrations took place. However,  
in practice, robots may also encounter unseen situations, such as obstacles or human intervention, which can be considered as new task constraints, requiring the robot to adapt its trajectory in order to perform adequately. 

In the context of imitation learning, several algorithms such as dynamic movement primitives (DMP) \citep{Ijspeert} and probabilistic movement primitives (ProMP) \citep{Paraschos} have been developed to generate desired trajectories in various scenarios. However, due to an explicit description of the trajectory dynamics, DMP introduces many open parameters in addition to basis functions and their weighting coefficients. The same problem arises in ProMP, which fits trajectories using basis functions that are manually defined. Moreover, DMP and ProMP were formulated towards the learning of time-driven trajectories (i.e., trajectories explicitly dependent on time), where the learning with high-dimensional inputs are not addressed.

In order to alleviate the modeling of trajectories via specific functions and meanwhile facilitate the learning of trajectories driven by high dimensional inputs, Gaussian mixture model (GMM) \citep{Calinon2007} has been employed to model the joint distribution of input variables and demonstrated motions. Usually, GMM is complemented with Gaussian mixture regression (GMR) \citep{Cohn} to retrieve a desired trajectory. Despite the improvements with respect to other techniques, adapting learned skills with GMM/GMR is not straightforward. Indeed, it is difficult to re-optimize GMM to fulfill new requirements (e.g., via-points), since this usually requires to re-estimate new model parameters (i.e., mixture coefficients, means and covariance matrices) that actually lie on a high-dimensional space . 

An alternative solution to refine trajectories for satisfying new task constraints is reinforcement learning (RL). For instance, a variant of policy improvement with path integrals \citep{Buchli} was employed to optimize the movement pattern of DMP \citep{Stulp}. Also, natural actor-critic \citep{Peters2005} was used to optimize the centers of GMM components \citep{Guenter}. However, the time-consuming search of the optimal policy might make the application of RL approaches to on-line refinements (such as those required after perturbations) impractical. In contrast to the RL treatment, ProMP formulates the modulation of trajectories as a Gaussian conditioning problem, and therefore derives an analytical solution to adapt trajectories towards new via-points or targets.
It is worth pointing out that DMP can adapt trajectories towards different goals, however, the via-points constraints are not addressed therein.

Besides the generation of adaptive trajectories, another desired property in imitation learning is extrapolation.
Often, human demonstrations are provided for a limited set of task instances, but the robot is expected to apply the learned movement patterns in a wider range of circumstances.
In this context, DMP is capable of generating trajectories starting from arbitrary locations and converging to a goal. This is achieved through a formulation based on a spring-damper system whose equilibrium corresponds to the target of the robot motion. In contrast, ProMP and GMM model the distribution of demonstrated trajectories in absolute frames rather than relative frames, which limits their extrapolation capabilities. As an extension of GMM, a task-parameterized formulation is studied in \cite{Calinon2016}, which in essence models local (or relative) trajectories and corresponding local patterns, therefore endowing GMM with better extrapolation performance.

While the aforementioned algorithms have achieved reliable performances, we aim for a solution that addresses the most crucial limitations of those approaches. In particular, we propose an algorithm that:
\begin{enumerate}[label=(\roman*)]
\item  preserves the probabilistic properties exhibited in multiple demonstrations, 
\item  deals with adaptation and superposition of trajectories,
\item  can be generalized for extrapolations, 
\item  learns human demonstrations associated with high-dimensional inputs while alleviating the need to explicitly define basis functions. 
\end{enumerate}	

The main contribution of this paper is the development of a novel \emph{kernelized movement primitive} (KMP), which allows us to address the above listed problems using a single framework. Specifically, KMP provides a non-parametric solution for imitation learning and hence alleviates the explicit representation of trajectories using basis functions, rendering fewer open parameters and easy implementation. More importantly, in light of the kernel treatment, KMP has the ability to model demonstrations associated with high-dimensional inputs, which is usually viewed as a non-trivial problem due to the curse of dimensionality. 

In addition, this paper extends KMP from a task-parameterized perspective and formulates \emph{local}-KMP, improving the extrapolation capabilities to different task situations described by a set of local coordinate frames. 
Finally, as a special case, we considers the application of KMP to the learning of time-driven trajectories, which inherits all the advantages of KMP while being suitable for time-scale modulation. For the sake of clear comparison, we list most relevant features of the state-of-the-art methods as well as our approach in Table~\ref{table:comp:table}. Note that we consider the modulation of robot trajectories to pass through desired via-points and end-points as the adaptation capability.

\begin{table}[bt]
	\caption {Comparison Among the State-of-the-Art and KMP}
	\centering
	\scalebox{0.9}{	
		\begin{tabular}{lcccc}
			\toprule %
			&DMP & ProMP & GMM & Our Approach\\ \toprule %
			$Probabilistic$ 
			& -
			& \checkmark 
			& \checkmark 
			& \checkmark 
			\\
			\midrule
			$Via\!\!-\!\!point$
			& -
			& \checkmark 
			& -
			& \checkmark
			\\
			\midrule
			$End\!\!-\!\!point$
			& \checkmark
			& \checkmark 
			& -
			& \checkmark
			\\
			\midrule
			$Extrapolation$ 
			& \checkmark
			& -
			& -
			& \checkmark
			\\
			\midrule
			$High$-$dim \; Inputs$
			& - 
			& -
			& \checkmark
			& \checkmark
			\\
			\bottomrule
		\end{tabular}}
		\label{table:comp:table}
	\end{table}

The structure of this paper is arranged as follows. 
We formulate imitation learning from an information-theory perspective and
propose KMP in Section~\ref{sec:kmp}. Subsequently, we extend KMP to deal with trajectory modulation and superposition in Section \ref{subsec:kmp:modulation} and Section~\ref{subsec:super:position}, respectively.  
Moreover, we introduce the concept of learning local trajectories into KMP in Section \ref{subsec:local_frame}, augmenting its extrapolation capabilities in task space. In Section~\ref{sec:time_kmp}, we discuss a special application of KMP to time-driven trajectories. We test the performance of KMP on trajectory modulation, superposition and extrapolation in Section \ref{sec:evaluations}, where
several scenarios are considered, ranging from learning handwritten letters to real robotic experiments.
After that, we review related work in Section~\ref{sec:relative:work}. An insightful discussion is provided in Section \ref{sec:discuss}, where we elaborate on the potential of our approach and the similarities between KMP and ProMP, as well as open challenges. Finally, we close with conclusions in Section~\ref{sec:conclusion}.

\section{Kernelized Representation of Movement Trajectory}
\label{sec:kmp}
Learning from multiple demonstrations allows for encoding trajectory distributions and extracting important or consistent features of the task. In this section, we first illustrate a probabilistic modeling of human demonstrations (Section~\ref{subsec:ref:traj}), and, subsequently, we exploit the resulting trajectory distribution to derive KMP (Section~\ref{subsec:kmp}). 

\subsection{Learning from Human Demonstrations}
\label{subsec:ref:traj}
Formally, let us denote the set of demonstrated training data by $\{\{ \vec{s}_{n,h},{\vec{\xi}}_{n,h}\}_{n=1}^{N}\}_{h=1}^{H}$ where $\vec{s}_{n,h} \in \mathbb{R}^{\mathcal{I}}$ is the input and ${\vec{\xi}}_{n,h} \in \mathbb{R}^{\mathcal{O}}$ denotes the output. Here, the super-indexes $\mathcal{I}$, $\mathcal{O}$, $H$ and $N$ respectively represent the dimensionality of the input and output space, the number of demonstrations, and the trajectory length. Note that a probabilistic encoding of the demonstrations allows the input $\vec{s}$ and output ${\vec{\xi}}$ to represent different types of variables. For instance, by considering $\vec{s}$ as the position of the robot and ${\vec{\xi}}$ as its velocity, the representation becomes an autonomous system. Alternatively, if $\vec{s}$ and ${\vec{\xi}}$ respectively represent time and position, the resulting encoding corresponds to a time-driven trajectory.   

In order to capture the probabilistic distribution of demonstrations, a number of algorithms can be employed, such as GMM \citep{Calinon2007}, hidden Markov models \citep{Leonel13}, and even a single Gaussian distribution \citep{Englert,Osa}, which differ in the type of information that is extracted from the demonstrations. As an example, let us exploit GMM as the model used to encode the training data. More specifically, GMM is employed to estimate the joint probability distribution $\mathcal{P}(\vec{s},\vec{\xi})$ from demonstrations, i.e., 
\begin{equation}
\left[\begin{matrix}
\vec{s}\\\vec{\xi}
\end{matrix}\right] \sim \sum_{c=1}^{C} \pi_c \mathcal{N}(\vec{\mu}_c,\vec{\Sigma}_c),
\label{equ:gmm}
\end{equation}
where $\pi_c$, $\vec{\mu}_c$ and $\vec{\Sigma}_c$ respectively represent the prior probability, mean and covariance of the $c$-th Gaussian component, while $C$ denotes the number of Gaussian components.

Furthermore, a probabilistic \emph{reference trajectory} $\{\hat{\vec{\xi}}_{n}\}_{n=1}^{N}$ can be retrieved via GMR \citep{Cohn,Calinon2016}, where each point ${\hat{\vec{\xi}}}_{n}$ associated with $\vec{s}_n$ is described by a conditional probability distribution with mean $\hat{\vec{\mu}}_n$ and covariance $\hat{\vec{\Sigma}}_n$, i.e, ${\hat{\vec{\xi}}}_{n}|\vec{s}_n\sim \mathcal{N}(\hat{\vec{\mu}}_{n},\hat{\vec{\Sigma}}_{n})$ (see Appendix~\ref{app:gmr} for details). 
This probabilistic reference trajectory encapsulates the variability in the demonstrations as well as the correlations among outputs. 
We take advantage of the probabilistic reference trajectory to derive KMP.

\subsection{Kernelized Movement Primitive (KMP)}
\label{subsec:kmp}

We start the derivation of KMP by considering a \emph{parametric trajectory} 
\begin{equation}
\vec{\xi}(\vec{s})
= \vec{\Theta}(\vec{s})^{\trsp} \vec{w}
\label{equ:linear:form}
\end{equation}
with the matrix $\vec{\Theta}(\vec{s})\in \mathbb{R}^{B\mathcal{O} \times \mathcal{O}}$ defined as follows
\begin{equation}
\vec{\Theta}(\vec{s})=\left[\begin{matrix} 
\vec{\varphi}(\vec{s}) & \vec{0} & \cdots &\vec{0} \\
\vec{0} & \vec{\varphi}(\vec{s}) &  \cdots &\vec{0} \\
\vdots & \vdots &  \ddots & \vdots \\
\vec{0} & \vec{0} &  \cdots & \vec{\varphi}(\vec{s})\\
\end{matrix}\right] ,
\label{equ:basis:function}
\end{equation}
and the weight vector $\vec{w} \in \mathbb{R}^{B\mathcal{O}}$, where $\vec{\varphi}(\vec{s})\in \mathbb{R}^{B}$ denotes $B$-dimensional basis functions\footnote{The treatment of fitting trajectories by using basis functions has also been studied in DMP \citep{Ijspeert} and ProMP \citep{Paraschos}.}.
Furthermore, we assume that the weight vector $\vec{w}$ is  normally distributed, i.e., ${\vec{w}\sim \mathcal{N}(\vec{\mu}_w,\vec{\Sigma}_w)}$, where the mean $\vec{\mu}_w$ and the covariance $\vec{\Sigma}_w$ are \emph{unknown}. Therefore, the parametric trajectory satisfies 
\begin{equation}
\vec{\xi}(\vec{s}) \sim \mathcal{N} \left(\vec{\Theta}(\vec{s})^{\trsp} \vec{\mu}_w, \vec{\Theta}(\vec{s})^{\trsp} \vec{\Sigma}_w \vec{\Theta}(\vec{s}) \right).
\label{equ:distribution:para:traj}
\end{equation} 
Note that our goal is to imitate the probabilistic reference trajectory $\{\hat{\vec{\xi}}_{n}\}_{n=1}^{N}$, thus we aim to match the parametric trajectory distribution formulated by (\ref{equ:distribution:para:traj}) with the reference trajectory distribution. In order to address this problem, we propose to minimize the \emph{Kullback-Leibler} divergence (KL-divergence) \citep{Kullback,Rasmussen} between both trajectory distributions (Section~\ref{subsubsec:kl}). Subsequently, we derive optimal solutions for both $\vec{\mu}_w$ and $\vec{\Sigma}_w$, and formulate KMP by using the kernel trick in Sections~\ref{subsubsec:optimal:mean:kmp} and \ref{subsubsec:optimal:var:kmp}, respectively.

\subsubsection{Imitation Learning Based on Information Theory:}
\label{subsubsec:kl}

Since the well-known KL-divergence can be used to measure the distance between two probability distributions, we here exploit it to optimize the parametric trajectory distribution so that it matches the reference trajectory distribution.
From the perspective of information transmission, the minimization of KL-divergence guarantees minimal information-loss in the process of imitation learning. 

Formally, we consider the minimization of the objective function
\begin{equation}
\begin{aligned}
J_{ini}(\vec{\mu}_w,\!\vec{\Sigma}_w)\!\!=\!\!\sum_{n=1}^{N}\!\! D_{KL} \biggl ( \mathcal{P}_{\mathbf{p}}(\vec{\xi}|\vec{s}_n)
|| \mathcal{P}_{\mathbf{r}} (\vec{\xi}|\vec{s}_n) \biggr),
\end{aligned}
\label{equ:kl:cost:ini:temp}
\end{equation}
where 
$\mathcal{P}_{\mathbf{p}}(\vec{\xi}|\vec{s}_n)$ represents the probability distribution of the parametric trajectory (\ref{equ:distribution:para:traj}) given the input $\vec{s}_n$, i.e.,
\begin{equation}
\mathcal{P}_{\mathbf{p}}(\vec{\xi}|\vec{s}_n)\!= \!\mathcal{N} \left(\vec{\xi}|\vec{\Theta}(\vec{s}_n)^{\trsp}\! \vec{\mu}_w, \vec{\Theta}(\vec{s}_n)^{\trsp} \vec{\Sigma}_w \vec{\Theta}(\vec{s}_n) \right),
\label{equ:def:prob:para}
\end{equation}
$\mathcal{P}_{\mathbf{r}}(\vec{\xi}|\vec{s}_n)$ corresponds to the probability distribution of the reference trajectory associated with $\vec{s}_n$ (as described in Section~\ref{subsec:ref:traj}), namely
\begin{equation}
\mathcal{P}_{\mathbf{r}}(\vec{\xi}|\vec{s}_n)=\mathcal{N} (\vec{\xi}|\hat{\vec{\mu}}_n,\hat{\vec{\Sigma}}_n).
\label{equ:def:prob:ref}
\end{equation}
$D_{KL}(\cdot||\cdot)$ denotes the KL-divergence between the probability distributions $\mathcal{P}_{\mathbf{p}}$ and $\mathcal{P}_{\mathbf{r}}$, which is defined by 
\begin{equation}
\begin{aligned}
D_{KL}(\mathcal{P}_{\mathbf{p}}&(\vec{\xi}|\vec{s}_n)||\mathcal{P}_{\mathbf{r}}(\vec{\xi}|\vec{s}_n))\\
&=\int \mathcal{P}_{\mathbf{p}}(\vec{\xi}|\vec{s}_n) \log \frac{\mathcal{P}_{\mathbf{p}}(\vec{\xi}|\vec{s}_n)}{\mathcal{P}_{\mathbf{r}}(\vec{\xi}|\vec{s}_n)} d\vec{\xi}.
\end{aligned}
\label{equ:kl:def}
\end{equation}
By using the properties of KL-divergence between two Gaussian distributions, we rewrite  (\ref{equ:kl:cost:ini:temp}) as 	
\begin{equation}
\begin{aligned}
J_{ini}(\vec{\mu}_w,\! \vec{\Sigma}_w)\! \! =\sum_{n=1}^{N}\! \frac{1}{2} \biggl ( 
\log|\hat{\vec{\Sigma}}_n|
\!-\!\log|\vec{\Theta}(\vec{s}_n)^{\trsp} \vec{\Sigma}_w \vec{\Theta}(\vec{s}_n)|\\
-\mathcal{O} 
+\mathrm{Tr}(\hat{\vec{\Sigma}}_n^{-1}
\vec{\Theta}(\vec{s}_n)^{\trsp} \vec{\Sigma}_w \vec{\Theta}(\vec{s}_n) )\\
+(\vec{\Theta}(\vec{s}_n)^{\trsp} \! \vec{\mu}_w \!\!-\!\!  \hat{\vec{\mu}}_{n})^{\trsp} \! \hat{\vec{\Sigma}}_n^{-1}\! (\vec{\Theta}(\vec{s}_n)^{\trsp} \! \vec{\mu}_w \!\!-\!\!  \hat{\vec{\mu}}_{n})
\biggr),
\end{aligned}
\label{equ:kl:cost:ini}
\end{equation}
where $| \, \cdot \, |$ and $\mathrm{Tr}(\cdot)$ denote the determinant and trace of a matrix, respectively. 

After removing the coefficient `$\frac{1}{2}$', the constant terms $\log|\hat{\vec{\Sigma}}_n|$ and $\mathcal{O}$, this objective function (\ref{equ:kl:cost:ini}) can be further decomposed into a \emph{mean minimization subproblem} and a \emph{covariance minimization subproblem}. The former is defined by minimizing
\begin{equation}
{J}_{ini}(\vec{\mu}_w)\!\!=\!\!\sum_{n=1}^{N} (\vec{\Theta}(\vec{s}_n)^{\trsp} \vec{\mu}_w \!- \!\hat{\vec{\mu}}_{n})^{\trsp} \hat{\vec{\Sigma}}_n^{-1} (\vec{\Theta}(\vec{s}_n)^{\trsp} \vec{\mu}_w \!- \! \hat{\vec{\mu}}_{n})
\label{equ:kl:mean:cost}
\end{equation}
and the latter is written as the minimization of
\begin{equation}
\begin{aligned}
{J}_{ini}(\vec{\Sigma}_w)=\sum_{n=1}^{N} \Big(&
-\log|\vec{\Theta}(\vec{s}_n)^{\trsp}  \vec{\Sigma}_w \vec{\Theta}(\vec{s}_n)|\\
&+\mathrm{Tr}(\hat{\vec{\Sigma}}_n^{-1}
\vec{\Theta}(\vec{s}_n)^{\trsp} \vec{\Sigma}_w \vec{\Theta}(\vec{s}_n)) \Big)
\end{aligned}.
\label{equ:kl:var:cost}
\end{equation}
In the following two sections,
we separately solve the mean and covariance subproblems, resulting in the KMP formulation.

\subsubsection{Mean Prediction of KMP:}
\label{subsubsec:optimal:mean:kmp}

In contrast to kernel ridge regression (KRR) \citep{Saunders, Murphy}, we introduce a penalty term $||\vec{\mu}_w||^2$ into the mean minimization subproblem (\ref{equ:kl:mean:cost}) so as to circumvent the over-fitting problem. Thus, the new mean minimization subproblem can be re-rewritten as
\begin{equation}
\begin{aligned}
	{J}(\vec{\mu}_w)\!\!=\!\!\!\sum_{n=1}^{N}\! (\vec{\Theta}(\vec{s}_n)^{\trsp}\!\vec{\mu}_w \!\!-\!\! \hat{\vec{\mu}}_{n})^{\trsp} \hat{\vec{\Sigma}}_n^{-1} (\vec{\Theta}&(\vec{s}_n)^{\trsp} \!\vec{\mu}_w \!\!-\!\! \hat{\vec{\mu}}_{n})\\
	&+\lambda \vec{\mu}_w^{\trsp}\vec{\mu}_w,
\end{aligned}
	\label{equ:kl:mean:cost:penalty}
\end{equation}
where $\lambda>0$.

The cost function (\ref{equ:kl:mean:cost:penalty}) resembles a weighted least squares formulation, except for the penalty term $\lambda \vec{\mu}_w^{\trsp}\vec{\mu}_w$. Also, it is similar to the common quadratic cost function minimized in KRR, where $\hat{\vec{\Sigma}}_n^{-1}=\vec{I}_{\mathcal{O}}$. However, the variability of the demonstrations encapsulated in $\hat{\vec{\Sigma}}_n$ is introduced in (\ref{equ:kl:mean:cost:penalty}) as an importance measure associated to each trajectory datapoint, which can be understood as relaxing or reinforcing the optimization for a particular datapoint. In other words, this covariance-weighted cost function permits large deviations from the reference trajectory points with high covariances, while demanding to be close when the associated covariance is low.

By taking advantage of the dual transformation of KRR, 
the optimal solution ${\vec{\mu}}_w^{*}$ of (\ref{equ:kl:mean:cost:penalty}) can be derived as (see \cite{Murphy,Kober2011} for details)
\begin{equation}
{\vec{\mu}}_w^{*}=\vec{\Phi} ( \vec{\Phi}^{\trsp} \vec{\Phi} +\lambda \vec{\Sigma} )^{-1} {\vec{\mu}},
\label{equ:kmp:muw}
\end{equation}
where 
\begin{equation}
\begin{aligned}
\vec{\Phi}&=[
\vec{\Theta}(\vec{s}_1) \ \vec{\Theta}(\vec{s}_2) \ \cdots \ \vec{\Theta}(\vec{s}_N)
],\\
\vec{\Sigma}&=\mathrm{blockdiag}(\hat{\vec{\Sigma}}_1, \ \hat{\vec{\Sigma}}_2, \ \ldots, \ \hat{\vec{\Sigma}}_N), \quad \\
{\vec{\mu}}&=[
\hat{\vec{\mu}}_1^{\trsp} \ \hat{\vec{\mu}}_2^{\trsp} \ \cdots \ \hat{\vec{\mu}}_N^{\trsp}
]^{\trsp}.
\end{aligned}
\label{equ:notations:define}
\end{equation}
Subsequently, 
for a query $\vec{s}^{*}$ (i.e., new input), its corresponding output (expected value) is computed as
\begin{equation}
\mathbb{E}(\vec{\xi}(\vec{s}^{*})) 
\!\!=\!\vec{\Theta}(\vec{s}^{*})^{\trsp}{\vec{\mu}}_w^{*}\!\!=\!\vec{\Theta}(\vec{s}^{*})^{\trsp}\vec{\Phi} ( \vec{\Phi}^{\trsp} \vec{\Phi} \!+\!\lambda \vec{\Sigma} )^{-1}  {\vec{\mu}}.
	\label{equ:kmp:mean:temp}
\end{equation}
In order to facilitate the application of (\ref{equ:kmp:mean:temp}) (particularly for high-dimensional $\vec{s}$), we propose to kernelize (\ref{equ:kmp:mean:temp}) so as to avoid the explicit definition of basis functions. Let us define the inner product for $\vec{\varphi}(\vec{s}_i)$ and $\vec{\varphi}(\vec{s}_j)$ as
\begin{equation}
\vec{\varphi}(\vec{s}_i)^{\trsp} \vec{\varphi}(\vec{s}_j)=k(\vec{s}_i,\vec{s}_j),
\label{equ:single:basis:product}
\end{equation} 
where $k(\cdot,\cdot)$ is a kernel function. Then, based on (\ref{equ:basis:function}) and (\ref{equ:single:basis:product}), we have 
\begin{equation}
\vec{\Theta}({\vec{s}_i})^{\trsp}\vec{\Theta}({\vec{s}_j})
=\left[\begin{matrix} 
k(\vec{s}_i, \vec{s}_j) & \vec{0} & \cdots &\vec{0} \\
\vec{0} & k(\vec{s}_i, \vec{s}_j) &  \cdots &\vec{0} \\
\vdots & \vdots &  \ddots & \vdots \\
\vec{0} & \vec{0} &  \cdots & k(\vec{s}_i, \vec{s}_j)\\
\end{matrix}\right],
\label{equ:basis:product}
\end{equation}
which can be further rewritten as 
a kernel matrix
\begin{equation}
	\vec{k}(\vec{s}_i, \ \vec{s}_j)= \vec{\Theta}({\vec{s}_i})^{\trsp}\vec{\Theta}({\vec{s}_j})=
	 k(\vec{s}_i, \vec{s}_j)\vec{I}_{\mathcal{O}},
	\label{equ:kernel:matrix}
\end{equation} 
where
$\vec{I}_{\mathcal{O}}$ is the ${\mathcal{O}}$-dimensional identity matrix.
Also, let us denote the matrix $\vec{{K}}$ as
\begin{equation}
\vec{K}
=\left[\begin{matrix} 
\vec{k}(\vec{s}_1, \vec{s}_1) & \vec{k}(\vec{s}_1, \vec{s}_2) & \cdots &\vec{k}(\vec{s}_1, \vec{s}_N) \\
\vec{k}(\vec{s}_2, \vec{s}_1) & \vec{k}(\vec{s}_2, \vec{s}_2) & \cdots &\vec{k}(\vec{s}_2, \vec{s}_N) \\
\vdots & \vdots &  \ddots & \vdots \\
\vec{k}(\vec{s}_N, \vec{s}_1) & \vec{k}(\vec{s}_N, \vec{s}_2) & \cdots &\vec{k}(\vec{s}_N, \vec{s}_N) \\
\end{matrix}\right],
\label{equ:K:matrix}
\end{equation}
and write the matrix $\vec{k}^{*}$ as
\begin{equation}
\vec{k}^{*}=[\vec{k}(\vec{s}^{*}, \vec{s}_{1}) \; \vec{k}(\vec{s}^{*}, \vec{s}_{2}) \; \cdots \; \vec{k}(\vec{s}^{*}, \vec{s}_{N})],
\label{equ:k:star}
\end{equation}
then the prediction in (\ref{equ:kmp:mean:temp}) becomes
\begin{equation}
\mathbb{E}(\vec{\xi}(\vec{s}^{*}))
=\ \vec{{k}}^{*} (\vec{{K}}+\lambda \vec{\Sigma})^{-1} {\vec{\mu}}.
	\label{equ:kmp:mean}
\end{equation}

Note that a similar result was derived in the context of reinforcement learning \citep{Kober2011} (called cost regularized kernel regression, CrKR). 
In contrast to the mean prediction of KMP, CrKR models target components separately without considering their correlations, i.e., a diagonal weighted matrix ${\vec{R}_n=r_n \vec{I}_{\mathcal{O}}}$ 
is used instead of the full covariance matrix $\hat{\vec{\Sigma}}_n^{-1}$ from (\ref{equ:kl:mean:cost:penalty}). Furthermore, for the case in which $\hat{\vec{\Sigma}}_n=\vec{I}_{\mathcal{O}}$, the prediction in (\ref{equ:kmp:mean}) is identical to the mean of the Gaussian process regression (GPR) \citep{Rasmussen}.

It is worth pointing out that the initial mean minimization subproblem (\ref{equ:kl:mean:cost}) is essentially equivalent to the problem of maximizing the posterior
$\prod_{n=1}^{N} \mathcal{P}(\vec{\Theta}(\vec{s}_n)^{\trsp} \vec{\mu}_w|\hat{\vec{\mu}}_n,\hat{\vec{\Sigma}}_n)$, please refer to Appendix~\ref{app:mean:dual} for the proof. Thus, the optimal solution ${\vec{\mu}}_w^{*}$ can be viewed as the best estimation given the observed reference trajectory distribution.

\subsubsection{Covariance Prediction of KMP:}
\label{subsubsec:optimal:var:kmp}

Similar to the treatment in (\ref{equ:kl:mean:cost:penalty}), we propose to add a penalty term into the covariance minimization subproblem (\ref{equ:kl:var:cost}) in order to bound the covariance $\vec{\Theta}(\vec{s}_n)^{\trsp} \vec{\Sigma}_w \vec{\Theta}(\vec{s}_n)$. On the basis of the properties of the \emph{Rayleigh quotient},
the penalty term could be defined by the largest eigenvalue of $\vec{\Sigma}_w$. For the sake of easy derivation, we impose a relaxed penalty term $\mathrm{Tr}(\vec{\Sigma}_w)$ which is larger than the largest eigenvalue of $\vec{\Sigma}_w$ since $\vec{\Sigma}_w$ is positive definite. 
Therefore, the new covariance minimization subproblem becomes
\begin{equation}
\begin{aligned}
{J}(\vec{\Sigma}_w)&=\sum_{n=1}^{N} \Big(-\log  |\vec{\Theta}(\vec{s}_n)^{\trsp} \vec{\Sigma}_w \vec{\Theta}(\vec{s}_n)|\\
&+\mathrm{Tr}(\hat{\vec{\Sigma}}_n^{-1}
\vec{\Theta}(\vec{s}_n)^{\trsp} \vec{\Sigma}_w \vec{\Theta}(\vec{s}_n)) \Big)+\lambda \mathrm{Tr}(\vec{\Sigma}_w)
\end{aligned}.
\label{equ:kl:var:cost:penalty}
\end{equation}

By computing the derivative of (\ref{equ:kl:var:cost:penalty}) with respect to $\vec{\Sigma}_w$
and setting it to 0, we have\footnote{The following results on matrix derivatives \citep{Petersen} are used:  $\frac{\partial |\vec{A}\vec{X}\vec{B}|}
	{\partial \vec{X}}=|\vec{A}\vec{X}\vec{B}|(\vec{X}^{\trsp})^{-1}$ and ${\frac{\partial}{\partial \vec{X}} \mathrm{Tr}(\vec{A}\vec{X}\vec{B})=\vec{A}^{\trsp} \vec{B}^{\trsp}}$.} 
\begin{equation}
\begin{aligned}
\sum_{n=1}^{N} \Big(
-\vec{\Sigma}_w^{-1}+ \vec{\Theta}(\vec{s}_n) \hat{\vec{\Sigma}}_n^{-1} \vec{\Theta}(\vec{s}_n)^{\trsp}
\Big) + \lambda \vec{I}=0
\end{aligned}.
\label{equ:kl:var:cost:derivative}
\end{equation}
Furthermore, we can rewrite (\ref{equ:kl:var:cost:derivative}) in a compact form by using $\vec{\Phi}$ and $\vec{\Sigma}$ from (\ref{equ:notations:define}) and derive the optimal solution $\vec{\Sigma}_w^{*}$ as follows
\begin{equation}
\vec{\Sigma}_w^{*}=N(\vec{\Phi}\vec{\Sigma}^{-1}\vec{\Phi}^{\trsp}+\lambda \vec{I})^{-1}.
\end{equation}
This solution resembles the covariance of weighted least square estimation, except for the factor `$N$' and the regularized term $\lambda \vec{I}$. 

According to the $\emph{Woodbury identity}$\footnote{$(\vec{A}+\vec{C}\vec{B}\vec{C}^{\trsp})^{-1}=\vec{A}^{-1}\!-\!\vec{A}^{-1}\vec{C}(\vec{B}^{-1}+\vec{C}^{\trsp}\vec{A}^{-1}\vec{C})^{-1}\vec{C}^{\trsp}\vec{A}^{-1}$.},
we can determine the covariance of $\vec{\xi}(\vec{s}^{*})$ for a query $\vec{s}^{*}$ as
\begin{equation}
\begin{aligned}
\mathbb{D}(\vec{\xi}(\vec{s}^{*}))&= \vec{\Theta}(\vec{s}^{*})^{\trsp}\vec{\Sigma}_w^{*} \vec{\Theta}(\vec{s}^{*})\\
&=N \vec{\Theta}(\vec{s}^{*})^{\trsp}  (\vec{\Phi}\vec{\Sigma}^{-1}\vec{\Phi}^{\trsp}+\lambda \vec{I})^{-1} \vec{\Theta}(\vec{s}^{*})\\
&=\frac{N}{\lambda} \vec{\Theta}(\vec{s}^{*})^{\trsp}  
\left(\vec{I} \!- \! \vec{\Phi}(\vec{\Phi}^{\trsp}\vec{\Phi}\!+\!\lambda\vec{\Sigma}）)^{-1}\vec{\Phi}^{\trsp} \right)
\vec{\Theta}(\vec{s}^{*}).
\end{aligned}
\label{equ:kmp:var:temp}
\end{equation}
Recall that we have defined the kernel matrix in (\ref{equ:kernel:matrix})-(\ref{equ:K:matrix}), and hence the covariance of $\vec{\xi}(\vec{s}^{*})$ becomes
\begin{equation}
\mathbb{D}(\vec{\xi}(\vec{s}^{*}))=\frac{N}{\lambda} \left(\vec{k}(\vec{s}^{*}, \vec{s}^{*}) -\vec{k}^{*}(\vec{K}+\lambda \vec{\Sigma})^{-1} \vec{k}^{*\trsp}\right).
\label{equ:kmp:var}
\end{equation}
In addition to the factor `$\frac{N}{\lambda}$', the covariance formula in (\ref{equ:kmp:var}) differs from the covariances defined in GPR and CrKR in two essential aspects. 
First, the variability $\vec{\Sigma}$ extracted from demonstrations (as defined in (\ref{equ:notations:define})) is used in the term $(\vec{K}+\lambda \vec{\Sigma})^{-1}$, while the identity matrix and the diagonal weighted matrix are used in GPR and CrKR, respectively. Second, in contrast to the diagonal covariances predicted by GPR and CrKR, KMP predicts a full matrix covariance which allows for predicting the correlations between output components. 
For the purpose of convenient descriptions in the following discussion, we refer to
${\vec{D}=\{\vec{s}_n,\hat{\vec{\mu}}_{n},\hat{\vec{\Sigma}}_{n}\}_{n=1}^{N}}$ as the \emph{reference database}. The prediction of both the mean and covariance using KMP is summarized in Algorithm~\ref{algorithm:kmp}.

\begin{algorithm}[t]
	\caption{\emph{Kernelized Movement Primitive}}
	\begin{algorithmic}[1]
\State{\textbf{{Initialization}}}
\Statex{- Define the kernel $k(\cdot,\cdot)$ and set the factor $\lambda$.} 
\State{\textbf{{Learning from demonstrations}}} (see Section~\ref{subsec:ref:traj})
\Statex{- Collect demonstrations $\{ \{ \vec{s}_{n,h},\vec{\xi}_{n,h}\}_{n=1}^{N} \}_{h=1}^{H}$}.
\Statex{- Extract the reference database $\{\vec{s}_n,\hat{\vec{\mu}}_{n},\hat{\vec{\Sigma}}_{n}\}_{n=1}^{N}$.}
\State{\textbf{{Prediction using KMP}}} (see Section~\ref{subsec:kmp})
\Statex{- {\emph{Input}}: query $\vec{s}^{*}$.}
\Statex{- Calculate $\vec{\Sigma}$, $\vec{\mu}$, $\vec{K}$ and $\vec{k}^{*}$ using (\ref{equ:notations:define}), (\ref{equ:K:matrix}) and (\ref{equ:k:star}).}
\Statex{- {\emph{Output}}: $\mathbb{E}(\vec{\xi}(\vec{s}^{*}))
=\ \vec{{k}}^{*} (\vec{{K}}+\lambda \vec{\Sigma})^{-1} {\vec{\mu}}$ \quad\,and
\Statex{$\mathbb{D}(\vec{\xi}(\vec{s}^{*}))=\frac{N}{\lambda} \left(\vec{k}(\vec{s}^{*}, \vec{s}^{*}) -\vec{k}^{*}(\vec{K}+\lambda \vec{\Sigma})^{-1} \vec{k}^{*\trsp}\right)$}. } 
	\end{algorithmic}
	\label{algorithm:kmp}
\end{algorithm}

\section{Extensions of Kernelized Movement Primitive} 
\label{sec:tp-kmp}
As previously explained, human demonstrations can be used to retrieve a distribution of trajectories that the robot exploits to carry out a specific task. However, in dynamic and unstructured environments the robot also needs to adapt its motions when required. 
For example, if an obstacle suddenly occupies an area that intersects the robot motion path, the robot is required to modulate its movement trajectory so that collisions are avoided. A similar modulation is necessary (e.g., in pick-and-place and reaching tasks) when the target varies its location during the task execution. The trajectory modulation problem will be addressed in Section~\ref{subsec:kmp:modulation} by exploiting the proposed KMP formulation.

Besides the modulation of a single trajectory, another challenging problem arises when the robot is given a set of candidate trajectories to follow, which represent feasible solutions for the task. Each of them may be assigned with a different priority (extracted, for example, from the task constraint). These candidate trajectories can be exploited to compute a mixed trajectory so as to balance all the feasible solutions according to their priorities. We cope with the superposition problem in Section~\ref{subsec:super:position} by using KMP.

Finally, human demonstrations are often provided in a relatively convenient task space. However, the robot might be expected to apply the learned skill to a broader domain. In order to address this problem, we extend KMP 
by using local coordinate systems and affine transformations as in \cite{Calinon2016}, which allows KMP to exhibit better extrapolation capabilities (Section \ref{subsec:local_frame}).

\subsection{Trajectory Modulation Using KMP}
\label{subsec:kmp:modulation}

We here consider trajectory modulation in terms of adapting trajectories to pass through new via-points/end-points.
Formally, let us define $M$ new desired points as $\{\bar{\vec{s}}_{m},\bar{\vec{\xi}}_m\}_{m=1}^{M}$ associated with conditional probability distributions $\bar{\vec{\xi}}_m | \bar{\vec{s}}_{m} \sim \mathcal{N} ( \bar{\vec{\mu}}_{m},\bar{\vec{\Sigma}}_{m} )$.
These conditional distributions can be designed based on new task requirements. For instance, if there are new via-points that the robot needs to pass through with high precision, small covariances $\bar{\vec{\Sigma}}_{m}$ are assigned. On the contrary, for via-points that allow for large tracking errors, high covariances can be set.

In order to consider both new desired points and the reference trajectory distribution simultaneously, we reformulate the original objective function defined in (\ref{equ:kl:cost:ini:temp}) as
\begin{equation}
\begin{aligned}
J_{ini}^{U}(\vec{\mu}_w,&\vec{\Sigma}_w)\!\!=\!\!
\sum_{n=1}^{N} \!D_{KL}\! \biggl (\! \mathcal{P}_{\mathbf{p}}(\vec{\xi}|\vec{s}_n)
|| \mathcal{P}_{\mathbf{r}} (\vec{\xi}|\vec{s}_n) \!\biggr)\\ &+\sum_{m=1}^{M}D_{KL} \biggl ( \mathcal{P}_{\mathbf{p}}(\vec{\xi}|\bar{\vec{s}}_m) || \mathcal{P}_{\mathbf{d}}(\vec{\xi}|\bar{\vec{s}}_m) \biggr)
\end{aligned}
\label{equ:kl:cost:ini:modulate}
\end{equation}
with
\begin{equation}
\mathcal{P}_{\mathbf{p}}(\vec{\xi}|\bar{\vec{s}}_m)\!\!=\!\!  \mathcal{N}\!\! \left(\vec{\xi}|\vec{\Theta}(\bar{\vec{s}}_m)^{\trsp}\! \vec{\mu}_w, \!\vec{\Theta}(\bar{\vec{s}}_m)^{\trsp} \vec{\Sigma}_w \vec{\Theta}(\bar{\vec{s}}_m) \right)
\label{equ:def:prob:para:des}
\end{equation}
and
\begin{equation}
\mathcal{P}_{\mathbf{d}}(\vec{\xi}|\bar{\vec{s}}_m)=\mathcal{N} (\vec{\xi}|\bar{\vec{\mu}}_m,\bar{\vec{\Sigma}}_m).
\label{equ:def:prob:ref:des}
\end{equation}
Let ${\bar{\vec{D}}=\{\bar{\vec{s}}_{m}, \bar{\vec{\mu}}_{m},\bar{\vec{\Sigma}}_{m}\}_{m=1}^{M}}$ denote the \emph{desired database}. We can concatenate the reference database $\vec{D}$ with the desired database $\bar{\vec{D}}$ and generate an \emph{extended reference database} $\{\vec{s}_{i}^{U},\vec{\mu}_{i}^{U},\vec{\Sigma}_{i}^{U}\}_{i=1}^{N+M}$, 
which is defined as follows
\begin{equation}
\left\{
\begin{aligned}
\vec{s}_{i}^{U}&\!=\!\vec{s}_{i}, \quad\,\, \vec{\mu}_{i}^{U}\!=\!\hat{\vec{\mu}}_{i}, \;\;\;\,\, \vec{\Sigma}_{i}^{U}\!=\!\hat{\vec{\Sigma}}_{i}, \;\;\,\,\,\,\, \mathrm{if} \;\;\; 1 \leq i \leq N\\
\vec{s}_{i}^{U}&\!=\!\bar{\vec{s}}_{i-N}, \vec{\mu}_{i}^{U}\!=\!\bar{\vec{\mu}}_{i-N}, \!\vec{\Sigma}_{i}^{U}\!=\!\bar{\vec{\Sigma}}_{i-N}, \mathrm{if}  \;N \!< i\!\leq\! N\!\!+\!\!M\\
\end{aligned}\right. \!,
\label{equ:combine:ref:desired}
\end{equation} 
Then, the objective function (\ref{equ:kl:cost:ini:modulate}) can be written as follows
\begin{equation}
\begin{aligned}
J_{ini}^{U}(\vec{\mu}_w,\!\vec{\Sigma}_w)\!\!=\!\!\!\sum_{i=1}^{M+N}\!\!\! D_{KL} \biggl (\! \mathcal{P}_{\mathbf{p}}(\vec{\xi}|\vec{s}_{i}^{U}) || \mathcal{P}_{\mathbf{u}}(\vec{\xi}|\vec{s}_{i}^{U}) \!\biggr),
\end{aligned}
\label{equ:kl:cost:ini:modulate:update:ref}
\end{equation}
with
\begin{equation}
\mathcal{P}_{\mathbf{p}}(\vec{\xi}|\vec{s}_{i}^{U})\!\!=\!\!  \mathcal{N}\!\! \left(\vec{\xi}|\vec{\Theta}(\vec{s}_{i}^{U})^{\trsp}\! \vec{\mu}_w, \!\vec{\Theta}(\vec{s}_{i}^{U})^{\trsp} \vec{\Sigma}_w \vec{\Theta}(\vec{s}_{i}^{U}) \right)
\label{equ:def:prob:para:extend}
\end{equation}
and
\begin{equation}
\mathcal{P}_{\mathbf{u}}(\vec{\xi}|\vec{s}_{i}^{U})=\mathcal{N} (\vec{\xi}|\vec{\mu}_i^{U},\vec{\Sigma}_i^{U}).
\label{equ:def:prob:ref:extend}
\end{equation}
Note that (\ref{equ:kl:cost:ini:modulate:update:ref}) has the same form as (\ref{equ:kl:cost:ini:temp}). Hence, for the problem of enforcing trajectories to pass through desired via-points/end-points, we can first concatenate the original reference database with the desired database through (\ref{equ:combine:ref:desired}) and, subsequently, with the extended reference database, we follow Algorithm~\ref{algorithm:kmp} to predict the mean and covariance for new queries $\vec{s}^{*}$.

It is worth pointing out that there might exist conflicts between the desired database and the original reference database. 
In order to illustrate this issue clearly, let us consider an extreme case: if there exist 
a new input $\bar{\vec{s}}_m=\vec{s}_n$, but $\bar{\vec{\mu}}_m$ is distant from $\hat{\vec{\mu}}_n$ while $\bar{\vec{\Sigma}}_m$ and $\hat{\vec{\Sigma}}_n$ are nearly the same, then the optimal solution of (\ref{equ:kl:cost:ini:modulate:update:ref}) corresponding to the query $\vec{s}_n$ can only be a trade-off between $\bar{\vec{\mu}}_m$ and $\hat{\vec{\mu}}_n$.
In the context of trajectory modulation using via-points/end-points, it is natural to consider new desired points with the highest preference. Thus, we propose to update the reference database from the perspective of reducing the above mentioned conflicts while maintaining most of datapoints in the reference database. 
The update procedure is carried out as follows.
For each datapoint $\{\bar{\vec{s}}_{m},\bar{\vec{\mu}}_m,\bar{\vec{\Sigma}}_m\}$ in the desired database, 
we first compare its input $\bar{\vec{s}}_{m}$ with the inputs $\{\vec{s}_{n}\}_{n=1}^{N}$ of the reference database so as to find the nearest datapoint $\{\vec{s}_{r},\hat{\vec{\mu}}_r,\hat{\vec{\Sigma}}_r\}$ that
satisfies ${d(\bar{\vec{s}}_m,\vec{s}_r) \leq d(\bar{\vec{s}}_m,\vec{s}_n), \forall n  \in  \{1,2,\ldots,N\}}$, where  $d(\cdot)$ could be an arbitrary distance measure such as 2-norm. 
If the nearest distance $d(\bar{\vec{s}}_m,\vec{s}_r)$ is smaller than a predefined threshold $\zeta>0$, we replace $\{\vec{s}_{r},\hat{\vec{\mu}}_r,\hat{\vec{\Sigma}}_r\}$ with $\{\bar{\vec{s}}_{m},\bar{\vec{\mu}}_m,\bar{\vec{\Sigma}}_m\}$; Otherwise, we insert $\{\bar{\vec{s}}_{m},\bar{\vec{\mu}}_m,\bar{\vec{\Sigma}}_m\}$ into the reference database. More specifically, given a new desired point $\{\bar{\vec{s}}_{m},\bar{\vec{\xi}}_m\}$ described by ${\bar{\vec{\xi}}_m | \bar{\vec{s}}_{m} \sim \mathcal{N} ( \bar{\vec{\mu}}_{m},\bar{\vec{\Sigma}}_{m} )}$, we update the reference database 
according to
\begin{equation}
	\left\{
	\begin{aligned}
		&\!\!\vec{{D}} \!\leftarrow\!  \{\! \vec{D}/\{\vec{s}_{r},\hat{\vec{\mu}}_{r},\hat{\vec{\Sigma}}_{r}\}\! \} \!\cup\! \{\bar{\vec{s}}_{m},\bar{\vec{\mu}}_{m},\bar{\vec{\Sigma}}_{m}\!\}, 
\;\mathrm{if} \, d(\bar{\vec{s}}_{m},\vec{s}_{r}) \!<\! \zeta,\\ 
		&\!\!\vec{{D}} \!\leftarrow   \vec{D} \cup \{\bar{\vec{s}}_{m},\bar{\vec{\mu}}_{m},\bar{\vec{\Sigma}}_{m}\}, 
		\hspace{0.85in} \mathrm{otherwise},
	\end{aligned}
	\right.
	\label{equ:kmp:update}
\end{equation}
where $r=\argmin_{n} d(\bar{\vec{s}}_{m},\vec{s}_{n}), n\in\{1,2,\ldots,N\}$ and the symbols `$/$' and `$\cup$' represent exclusion and union operations, respectively.

\subsection{Trajectory Superposition Using KMP}
\label{subsec:super:position}
In addition to the modulation operations on a single trajectory, we extend KMP to mix multiple trajectories that represent different feasible solutions for a task,  with different priorities. Formally, given a set of $L$ reference trajectory distributions, associated with inputs and corresponding priorities $\gamma_{n,l}$, denoted as $\{ \{\vec{s}_{n},\hat{\vec{\xi}}_{n,l},\gamma_{n,l}\}_{n=1}^{N}\}_{l=1}^{L}$, where ${\hat{\vec{\xi}}_{n,l}|\vec{s}_n \sim \mathcal{N}(\hat{\vec{\mu}}_{n,l},\hat{\vec{\Sigma}}_{n,l})}$, and $\gamma_{n,l} \in (0,1)$ is a priority assigned to the point $\{\vec{s}_{n},\hat{\vec{\xi}}_{n,l}\}$ satisfying $\sum_{l=1}^{L}\gamma_{n,l}=1$.

Since each priority indicates the importance of one datapoint in a reference trajectory, we use them to weigh the information-loss as follows
\begin{equation}
\begin{aligned}
J_{ini}^{S}(\vec{\mu}_w,\!\vec{\Sigma}_w)\!\!=\!\!\sum_{n=1}^{N}\!\sum_{l=1}^{L}\!\!\gamma_{n,l} D_{KL} \biggl (\!\! \mathcal{P}_{\mathbf{p}}(&\vec{\xi}|\vec{s}_n)
|| \mathcal{P}^{l}_{\mathbf{s}}(\vec{\xi}|\vec{s}_n) \!\!\biggr),
\end{aligned}
\label{equ:kl:cost:ini:mix}
\end{equation}
where $\mathcal{P}^{l}_{\mathbf{s}}$ is defined as
\begin{equation}
\mathcal{P}^{l}_{\mathbf{s}}(\vec{\xi}|\vec{s}_n)=\mathcal{N} (\vec{\xi}|\hat{\vec{\mu}}_{n,l},\hat{\vec{\Sigma}}_{n,l}),
\label{equ:def:prob:subref}
\end{equation}
representing the distribution of the $l$-th reference trajectory given the input $\vec{s}_n$.

Similar to the decomposition in (\ref{equ:kl:cost:ini})--(\ref{equ:kl:var:cost}), the objective function (\ref{equ:kl:cost:ini:mix}) can be decomposed into a \emph{weighted mean minimization subproblem} and a \emph{weighted covariance minimization subproblem}. The former is written as
\begin{equation}
\begin{aligned}
{J}_{ini}^{S}(\vec{\mu}_w)\!\!=\!\!\sum_{n=1}^{N}\! \sum_{l=1}^{L}  \gamma_{n,l}(\vec{\Theta}(\vec{s}_n)^{\trsp} &\vec{\mu}_w -\hat{\vec{\mu}}_{n,l})^{\trsp} \hat{\vec{\Sigma}}_{n,l}^{-1} \\
&(\vec{\Theta}(\vec{s}_n)^{\trsp} \vec{\mu}_w \!-\! \hat{\vec{\mu}}_{n,l})
\end{aligned}
\label{equ:kl:mean:cost:mix}
\end{equation}
and the latter is
\begin{equation}
\begin{aligned}
{J}_{ini}^{S}(\vec{\Sigma}_w)\!=\!\sum_{n=1}^{N} &\sum_{l=1}^{L}
\gamma_{n,l}\Big(\!-\!\log|\vec{\Theta}(\vec{s}_n)^{\trsp}  \vec{\Sigma}_w \vec{\Theta}(\vec{s}_n)|\\
&+\mathrm{Tr}(\hat{\vec{\Sigma}}_{n,l}^{-1}
\vec{\Theta}(\vec{s}_n)^{\trsp} \vec{\Sigma}_w \vec{\Theta}(\vec{s}_n)) \Big)
\end{aligned}.
\label{equ:kl:var:cost:mix}
\end{equation}
 
It can be proved that the weighted mean subproblem can be solved by minimizing (see Appendix~\ref{app:compose:mean})
\begin{equation}
\tilde{J}_{ini}^{S}(\vec{\mu}_w)\!\!=\!\!\sum_{n=1}^{N} (\vec{\Theta}(\vec{s}_n)^{\trsp} \!\vec{\mu}_w \!-\! \vec{\mu}_{n}^{S})^{\trsp} {\vec{\Sigma}_n^{S}}^{-1} (\vec{\Theta}(\vec{s}_n)^{\trsp} \!\vec{\mu}_w \!-\! \vec{\mu}_{n}^{S})
\label{equ:kl:mean:cost:mix:prod}
\end{equation}
and the weighted covariance subproblem is equivalent to the problem of minimizing (see Appendix~\ref{app:compose:var})
\begin{equation}
\begin{aligned}
\tilde{J}_{ini}^{S}(\vec{\Sigma}_w)\!=\!\sum_{n=1}^{N} \Big(&
\!\!-\log |\vec{\Theta}(\vec{s}_n)^{\trsp}  \vec{\Sigma}_w \vec{\Theta}(\vec{s}_n)|\\
&+\mathrm{Tr}({\vec{\Sigma}_n^{S}}^{-1}
\vec{\Theta}(\vec{s}_n)^{\trsp} \vec{\Sigma}_w \vec{\Theta}(\vec{s}_n)) \Big)
\end{aligned},
\label{equ:kl:var:cost:mix:prod}
\end{equation}
where
\begin{equation}
{\vec{\Sigma}_n^{S}}^{-1}=\sum_{l=1}^{L} \left( \hat{\vec{\Sigma}}_{n,l}/\gamma_{n,l} \right)^{-1} \quad \mathrm{and}
\label{equ:prod:var}
\end{equation}
\begin{equation}
\vec{\mu}_{n}^{S}={\vec{\Sigma}_n^{S}} \sum_{l=1}^{L} \left( \hat{\vec{\Sigma}}_{n,l}/\gamma_{n,l} \right)^{-1}  \hat{\vec{\mu}}_{n,l}.
\label{equ:prod:mean}
\end{equation} 
Observe that (\ref{equ:kl:mean:cost:mix:prod}) and (\ref{equ:kl:var:cost:mix:prod}) have the same form as the subproblems defined in (\ref{equ:kl:mean:cost}) and (\ref{equ:kl:var:cost}), respectively. Note that the definitions in (\ref{equ:prod:var}) and (\ref{equ:prod:mean}) essentially correspond to the product of $L$ Gaussian distributions $\mathcal{N}(\hat{\vec{\mu}}_{n,l},\hat{\vec{\Sigma}}_{n,l}/\gamma_{n,l})$ with ${l=1,2,\ldots,L}$, given by
\begin{equation}
\mathcal{N}(\vec{\mu}_{n}^{S},{\vec{\Sigma}_n^{S}}) \propto \prod_{l=1}^{L} \mathcal{N}(\hat{\vec{\mu}}_{n,l},\hat{\vec{\Sigma}}_{n,l}/\gamma_{n,l}).
\label{equ:product:mix}
\end{equation}
Thus, for the problem of trajectory superposition, we first determine a \emph{mixed reference database}
$\{\vec{s}_n,\vec{\mu}_n^{S},\vec{\Sigma}_n^{S}\}_{n=1}^{N}$ through (\ref{equ:product:mix}), 
then we employ Algorithm~\ref{algorithm:kmp} to predict the corresponding mixed trajectory points for arbitrary queries. 
Note that the weighted mean minimization subproblem (\ref{equ:kl:mean:cost:mix}) can be interpreted as the maximization of the weighted posterior 
$\prod_{n=1}^{N} \prod_{l=1}^{L} \mathcal{P} \left( \vec{\Theta}(\vec{s}_{n})^{\trsp} \vec{\mu}_w|\hat{\vec{\mu}}_{n,l},\hat{\vec{\Sigma}}_{n,l} \right)^{\gamma_{n,l}}.$
In comparison with the trajectory mixture in ProMP \citep{Paraschos}, 
we here consider an optimization problem with an unknown $\vec{\mu}_w$ rather than the direct product of a set of known probabilities.

\subsection{Local Movement Learning Using KMP}
\label{subsec:local_frame}

\begin{algorithm}[t]
	\caption{\emph{Local Kernelized Movement Primitives with Via-points/End-points}}
	\begin{algorithmic}[1]
	\State{\textbf{{Initialization}}}
		\Statex{- Define $k(\cdot,\cdot)$ and set $\lambda$.} 
		\Statex{- Determine $P$ local frames $\{\vec{A}^{(p)},\vec{b}^{(p)}\}_{p=1}^{P}$.}
	\State{\textbf{{Learning from local demonstrations}}}
		\Statex{- Collect demonstrations $\{ \{ \vec{s}_{n,h},\vec{\xi}_{n,h}\}_{n=1}^{N} \}_{h=1}^{H}$ in $\{O\}$.}
		\Statex{- Project demonstrations into local frames via (\ref{equ:linear:transform}).}
		\Statex{- Extract local reference databases $\!\{\vec{s}_n^{(p)}\!,\!\hat{\vec{\mu}}_n^{(p)}\!,\!\hat{\vec{\Sigma}}_n^{(p)}\}_{n=1}^{N}\!$.}
	\State{\textbf{{Update local reference databases}}}
		\Statex{- Project via-points/end-points into local frames via (\ref{equ:linear:transform}).} 
		\Statex{- Update local reference databases via (\ref{equ:kmp:update}).}
		\Statex{- Update $\vec{K}^{(p)},\vec{\mu}^{(p)},\vec{\Sigma}^{(p)}$ in each frame $\{p\}$.}
	\State{\textbf{{Prediction using local-KMPs}}} 
		\Statex{- {\emph{Input}}: query $\vec{s}^{*}$.}
		\Statex{- Update $P$ local frames based on new task requirements.} 
		\Statex{- Project $\vec{s}^{*}$ into local frames via (\ref{equ:linear:transform}), yielding $\{\!\vec{s}\!^{*\!(p)}\!\}\!_{p=1}^{P}$.}
		\Statex{- Predict the local trajectory point associated with $\vec{s}^{*(p)}$ in each frame $\{p\}$ using KMP.} 
		\Statex{- {\emph{Output}}: Compute $\widetilde{\vec{\xi}}(\vec{s}^{*})$ in the frame $\{O\}$ using (\ref{equ:local:product}).}
	\end{algorithmic}
	\label{algorithm:local-kmp}
\end{algorithm}

So far we have considered trajectories that are represented with respect to the same global frame (coordinate system).
In order to enhance the extrapolation capability of KMP in task space, human demonstrations can be encoded in local frames\footnote{Also referred to as task parameters in \cite{Calinon2016}.} so as to extract local movement patterns, which can then be applied to a wider range of task instances. 
Usually, the definition of local frames depends on the task at hand.   
For example, in a transportation task where the robot moves an object from a starting position (that may vary) to different target locations, two local frames can be defined respectively at the starting and ending positions.

Formally, let us define $P$ local frames as $\{\vec{A}^{(p)},\vec{b}^{(p)}\}_{p=1}^{P}$,
where $\vec{A}^{(p)}$ and $\vec{b}^{(p)}$ respectively represent the rotation matrix and the translation vector of frame $\{p\}$ with respect to the base frame $\{O\}$. Demonstrations are projected into each frame $\{p\}$, resulting in new trajectory points $\{ \{\vec{s}_{n,h}^{(p)},\vec{\xi}_{n,h}^{(p)}\}_{n=1}^{N} \}_{h=1}^{H}$ for each local frame, where  
\begin{equation}
	\left[ \begin{matrix}
		\vec{s}_{n,h}^{(p)} \\ \vec{\xi}_{n,h}^{(p)} 
	\end{matrix} \right]=
	\left[\begin{matrix}
	\vec{A}_{s}^{(p)} &\vec{0}\\
	\vec{0}         & \vec{A}_{\xi}^{(p)}	
	\end{matrix}\right]^{-1} 
	\left( 
	\left[ \begin{matrix}
		\vec{s}_{n,h} \\ \vec{\xi}_{n,h}
	\end{matrix} \right]-
	\left[\begin{matrix}
	\vec{b}_{s}^{(p)} \\
	\vec{b}_{\xi}^{(p)}	
	\end{matrix}\right] 
	\right) ,
	\label{equ:linear:transform}
\end{equation}
with $\vec{A}_{s}^{(p)}\!=\vec{A}_{\xi}^{(p)}\!=\vec{A}^{(p)}$ and $\vec{b}_{s}^{(p)}\!=\vec{b}_{\xi}^{(p)}\!=\vec{b}^{(p)}$\footnote{Note that, if the input $\vec{s}$ becomes time, then $\vec{A}_{s}^{(p)}=\!1$ and $\vec{b}_{s}^{(p)}=\!0$.}.
Subsequently, by following the procedure in Section~\ref{subsec:ref:traj}, for each local frame $\{p\}$ we can generate 
a \emph{local reference database} ${\vec{D}^{(p)}=\{\vec{s}_n^{(p)},\hat{\vec{\mu}}^{(p)}_{n},\hat{\vec{\Sigma}}^{(p)}_{n}\}_{n=1}^{N}}$.

We refer to the learning of KMPs in local frames as \emph{local}-KMPs. For sake of simplicity, we only discuss the trajectory modulations with via-points/end-points. The operation of trajectory superposition can be treated in the similar manner.
Given a set of desired points in the robot base frame $\{O\}$ described by the desired database $\{\bar{\vec{s}}_{m},\bar{\vec{\mu}}_m,\bar{\vec{\Sigma}}_m\}_{m=1}^{M}$, we project the desired database into local frames using (\ref{equ:linear:transform}), leading to the set of transformed \emph{local desired databases} $\bar{\vec{D}}^{(p)}=\{\bar{\vec{s}}_{m}^{(p)},\bar{\vec{\mu}}_m^{(p)},\bar{\vec{\Sigma}}_m^{(p)}\}_{m=1}^{M}$ with $p=\{1,2,\dots,P\}$. Then, we carry out the update procedure described by (\ref{equ:kmp:update}) in each frame $\{p\}$ and obtain a new local reference database $\vec{D}^{(p)}$.  

For a new input $\vec{s}^{*}$ in the base frame $\{O\}$, we first project it into local frames using the input transformation in (\ref{equ:linear:transform}), yielding local inputs $\{\vec{s}^{*(p)}\}_{p=1}^{P}$. Note that, during the prediction phase, local frames might be updated depending on new task requirements and  the corresponding task parameters $\vec{A}^{(p)}$ and $\vec{b}^{(p)}$ might vary accordingly. Later, in each frame $\{p\}$ we can predict a local trajectory point $\widetilde{\vec{\xi}}^{(p)}\!\!( \vec{s}^{*(p)})\sim \mathcal{N}( {\vec{\mu}}^{*(p)} , {\vec{\Sigma}}^{*(p)} )$ with updated mean ${\vec{\mu}}^{*(p)}$ and covariance ${\vec{\Sigma}}^{*(p)}$ by using (\ref{equ:kmp:mean}) and (\ref{equ:kmp:var}).  
Furthermore, new local trajectory points from all local frames can be simultaneously transformed into the robot base frame using an inverse formulation of (\ref{equ:linear:transform}). Thus, for the query  $\vec{s}^{*}$ in the base frame $\{O\}$, its corresponding trajectory point $\widetilde{\vec{\xi}}(\vec{s}^{*})$ in $\{O\}$ can be determined by maximizing the product of linearly transformed Gaussian distributions
\begin{equation}
\widetilde{\vec{\xi}}(\vec{s}^{*})\!=\!\argmax_{\vec{\xi}}\!\prod_{p=1}^{P}\! \mathcal{N}\! \biggl(\! \vec{\xi} | \underbrace{\vec{A}_{\xi}^{(p)} \vec{{\mu}}^{*(p)} \!\!+\! \vec{b}_{\xi}^{(p)}}_{\widetilde{\vec{\mu}}_p}, \underbrace{\vec{A}_{\xi}^{(p)} {\vec{\Sigma}}^{*(p)} \!\!{\vec{A}_{\xi}^{(p)}}^{\trsp}}_{\widetilde{\vec{\Sigma}}_p} \biggr)\!, 
\label{equ:local:to:global}
\end{equation}
whose optimal solution is
\begin{equation}
\widetilde{\vec{\xi}}(\vec{s}^{*})=\biggl(\sum_{p=1}^{P} \widetilde{\vec{\Sigma}}_p^{-1} \biggr)^{-1} \sum_{p=1}^{P} \widetilde{\vec{\Sigma}}_p^{-1} \widetilde{\vec{\mu}}_p.
\label{equ:local:product}
\end{equation}
The described procedure is summarized in Algorithm \ref{algorithm:local-kmp}. 
Note that the solution (\ref{equ:local:product}) actually corresponds to the expectation part of the product of Gaussian distributions in (\ref{equ:local:to:global}).

\begin{figure*}[bt] \centering
	\subfigure[]{ 
		\includegraphics[width=0.32\textwidth]{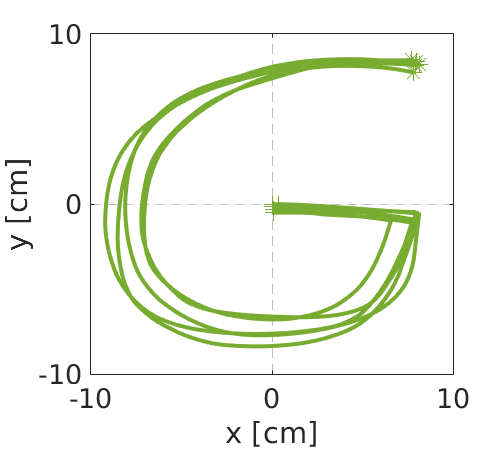}}  
	\subfigure[]{ 
		\includegraphics[width=0.32\textwidth]{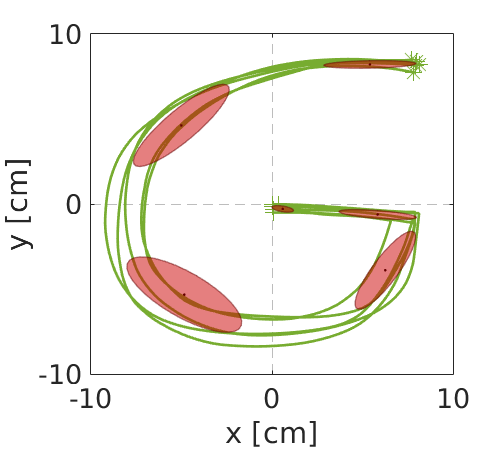}} 
	\subfigure[]{ 
		\includegraphics[width=0.32\textwidth]{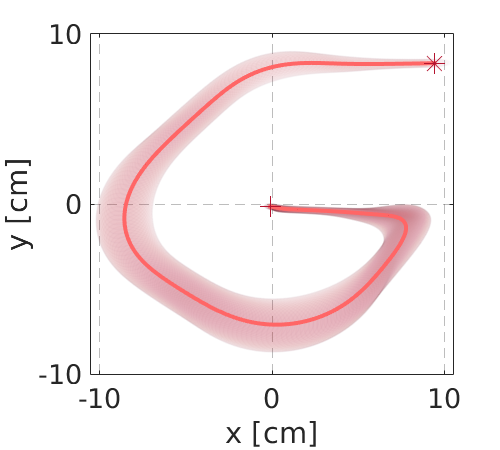}} 
	\caption{Demonstrations of handwritten letter `G' and the estimation of the reference trajectory through GMM/GMR. (\emph{a}) shows the trajectories of `G', where `$\ast$' and `+' denote the starting and ending points of the demonstrations, respectively. 
		(\emph{b}) depicts the estimated GMM with the ellipses representing Gaussian components. (\emph{c}) displays the retrieval of the reference trajectory distribution, where the red solid curve and shaded area, respectively, correspond to the mean and standard deviation of the reference trajectory.} 
	\label{fig:g:demos} 
\end{figure*}

\section{Time-driven Kernelized Movement Primitives}
\label{sec:time_kmp}

In many robotic tasks, such as biped locomotion \citep{Nakanishi} and striking movements \citep{Huang2016}, time plays a critical role when generating movement trajectories for a robot. We here consider a special case of KMP by taking time $t$ as the input $\vec{s}$, which is aimed at learning time-driven trajectories.

\subsection{A Special Treatment of Time-Driven KMP}
\label{subsec:time:tmp}
Similarly to ProMP, we formulate a parametric trajectory comprising positions and velocities as
\begin{equation}
\left[ \begin{matrix}
\vec{\xi}(t) \\ \dot{\vec{\xi}}(t) 
\end{matrix} \right] = \vec{\Theta}(t)^{\trsp} \vec{w},
\label{equ:linear:form:time}
\end{equation}
where the matrix $\vec{\Theta}(t)\in \mathbb{R}^{B\mathcal{O} \times 2\mathcal{O}}$ is
\begin{equation}
\vec{\Theta}(t)\!\!=\!\!\left[\begin{matrix} 
\vec{\varphi}(t) \!& \vec{0} \!& \cdots \!&\vec{0} \!& \dot{\vec{\varphi}}(t) \!& \vec{0} \!& \cdots \!&\vec{0} \\
\vec{0} \!& \vec{\varphi}(t) \!&  \cdots \!&\vec{0} \!&\vec{0} \!& \dot{\vec{\varphi}}(t) \!&  \cdots &\vec{0}\\
\vdots \!& \vdots \!&  \ddots \!& \vdots \!&\vdots \!& \vdots \!&  \ddots \!& \vdots\\
\vec{0} \!& \vec{0} \!&  \cdots \!& \vec{\varphi}(t) \!&\vec{0} \!& \vec{0} \!&  \cdots \!& \dot{\vec{\varphi}}(t)\\
\end{matrix}\right] \!.
\label{equ:basis:function:time}
\end{equation}
Note that we have included the first-order derivative of the parametric trajectory $\vec{\xi}(t)$ in (\ref{equ:linear:form:time}), which allows us to encode the observed dynamics of the motion. Consequently, we include the derivative of basis functions as shown in (\ref{equ:basis:function:time})

In order to encapsulate the variability in demonstrations, we here model the joint probability $\mathcal{P}(t,\vec{\xi},\dot{\vec{\xi}})$ using GMM, similarly to Section~\ref{subsec:ref:traj}. The probabilistic reference trajectory associated with time input $t_n$ can then be extracted by GMR as the conditional probability 
${\mathcal{P}(\hat{\vec{\xi}}_n,\hat{\dot{\vec{\xi}}}_n|t_n)
\sim \mathcal{N}(\hat{\vec{\mu}}_{n},\hat{\vec{\Sigma}}_{n})}$. 
Finally, we can derive the time-driven KMP by following the derivations presented in Section~\ref{subsec:kmp}. 

It is noted that, when we calculate the kernel matrix as previously defined in (\ref{equ:single:basis:product})--(\ref{equ:kernel:matrix}), we here encounter four types of products $\vec{\varphi}(t_i)^{\trsp} \vec{\varphi}(t_j)$, $\vec{\varphi}(t_i)^{\trsp} \dot{\vec{\varphi}}(t_j)$, $\dot{\vec{\varphi}}(t_i)^{\trsp} \vec{\varphi}(t_j)$ and $\dot{\vec{\varphi}}(t_i)^{\trsp} \dot{\vec{\varphi}}(t_j)$. 
Hence, we propose to approximate $\vec{\dot{\varphi}}(t)$ as
$\vec{\dot{\varphi}}(t) \approx \frac{\vec{\varphi}(t+\delta)-\vec{\varphi}(t)}{\delta}$ by using the finite difference method, where $\delta>0$ is an extremely small constant. So, based on the definition $\vec{\varphi}(t_i)^{\trsp} \vec{\varphi}(t_j)=k(t_i,t_j)$, we can determine the kernel matrix as
\begin{equation}
\vec{k}(t_i,t_j)\!=\! \vec{\Theta}({t_i})^{\trsp}\vec{\Theta}({t_j})\!\!=\!\!
\left[ \begin{matrix} k_{tt}(i,j)\vec{I}_{\mathcal{O}} \!&\! k_{td}(i,j)\vec{I}_{\mathcal{O}}\\
k_{dt}(i,j)\vec{I}_{\mathcal{O}} \!&\! k_{dd}(i,j)\vec{I}_{\mathcal{O}} \\
\end{matrix} \right],
\label{equ:kernel:matrix:time}
\end{equation} 
where
\begin{equation}
\begin{aligned}
k_{tt}(i,j)\!\!&=\!\!k(t_i,t_j),\\
k_{td}(i,j)\!\!&=\!\!\frac{k(t_i,t_j\!+\!\delta)\!\!-\!\!k(t_i,t_j)}{\delta},\\
k_{dt}(i,j)\!\!&=\!\!\frac{k(t_i\!+\!\delta,t_j)\!\!-\!\!k(t_i,t_j)}{\delta}, \\
k_{dd}(i,j)\!\!&=\!\!\frac{k(t_i\!+\!\delta, \!t_j\!+\!\delta)\!\! -\!\!k(t_i\!+\!\delta, \!t_j)\!\! -\!\!k(t_i,\!t_j\!+\!\delta)\!\! +\!\!k(t_i,\!t_j)}{{\delta}^{2}}. 
\end{aligned}
\end{equation}
It follows that we can actually model the output variable $\vec{\xi}(t)$ and its derivative $\dot{\vec{\xi}}(t)$ in (\ref{equ:linear:form:time}) using 
${\vec{\Theta}(t)=blockdiag(
\vec{\varphi}(t),\vec{\varphi}(t),\cdots,\vec{\varphi}(t))}$.
In other words, the derivative of basis functions is not used. However, this treatment requires a higher dimensional $\vec{\Theta}(t)$, (i.e., $2B\mathcal{O}\times 2\mathcal{O}$) and thus leads to a higher dimensional $\vec{w}\in\mathbb{R}^{2B\mathcal{O}}$. In contrast, if both basis functions and their derivatives (as defined in (\ref{equ:basis:function:time})) are employed, we can obtain a compact representation which essentially corresponds to a lower dimensional $\vec{w}\in\mathbb{R}^{B\mathcal{O}}$.

While the derivation presented in this section applies for a time-driven case, it cannot be easily generalized to the case of high-dimensional $\vec{s}$. Unlike a straightforward approximation of $\vec{\dot{\varphi}}(t)$ by using the finite difference method, for the high-dimensional input $\vec{s}$ it is a non-trivial problem to estimate $\vec{\dot{\varphi}}(s)=\frac{\partial \vec{\varphi}(s)}{\partial s}\frac{\partial s}{\partial t}$
unless we have an additional model which can reflect the dynamics between time $t$ and the input $\vec{s}$. Due to the difficulty of estimating $\vec{\dot{\varphi}}(s)$, an alternative way to encode $[\vec{\xi}^{\trsp}(\vec{s}) \  \dot{\vec{\xi}}^{\trsp}(\vec{s})]^{\trsp}$ with high-dimensional input $\vec{s}$ is to use (\ref{equ:linear:form}) with an extended matrix $\vec{\Theta}(\vec{s}) \in \mathbb{R}^{2B\mathcal{O} \times 2\mathcal{O}}$, i.e., ${\vec{\Theta}(\vec{s})=blockdiag(
	\vec{\varphi}(\vec{s}),\vec{\varphi}(\vec{s}),\cdots,\vec{\varphi}(\vec{s}))}$.

\subsection{Time-scale Modulation of Time-driven KMP}
\label{subsec:time:modulation}
In the context of time-driven trajectories, new tasks may demand to speed up or slow down the robot movement, and hence the trajectory modulation on the time-scale is required. Let us denote the movement duration of demonstrations and the time length of the corresponding reference trajectory as $t_N$. To generate adapted trajectories with new durations $t_D$, we define a monotonic function ${\tau: [0,t_D] \mapsto [0,t_N]}$, which is a transformation of time.
This straightforward solution implies that for any query $t^{*}\in[0,t_D]$, we use $\tau(t^{*})$ as the input for the prediction through KMP, and thus trajectories can be modulated as faster or slower (see also \cite{Ijspeert, Paraschos} for the modulations in time-scale, where the time modulation is called the phase transformation.).

\section{Evaluations of the Approach}
\label{sec:evaluations}
In this section, several examples are used to evaluate KMP. We first consider the adaptation of trajectories  with via-points/end-points as well as the mixture of multiple trajectories (Section~\ref{subsec:traj:modulate}), where comparisons with ProMP are shown. Then, we evaluate the extrapolation capabilities of local-KMPs (Section~\ref{subsec:extra:evaluate}). Subsequently, we validate the approach in two different scenarios using real robots. First, we study a novel application of robot motion adaptation by adding via-points according to sensed forces at the end-effector of the robot (Section~\ref{subsec:force:ada}). Second, we focus on a human-robot collaboration scenario, namely, the 3rd-hand task, where a 6-dimensional input is considered in the learning and adaptation problems (Section~\ref{subsec:3rd:hand}).

\begin{figure*} \centering 
	\subfigure[Trajectory modulations with one start-point and one via-point.]{
		\includegraphics[width=0.80\textwidth,bb=0.0cm 9.2cm 29cm 24cm,clip]{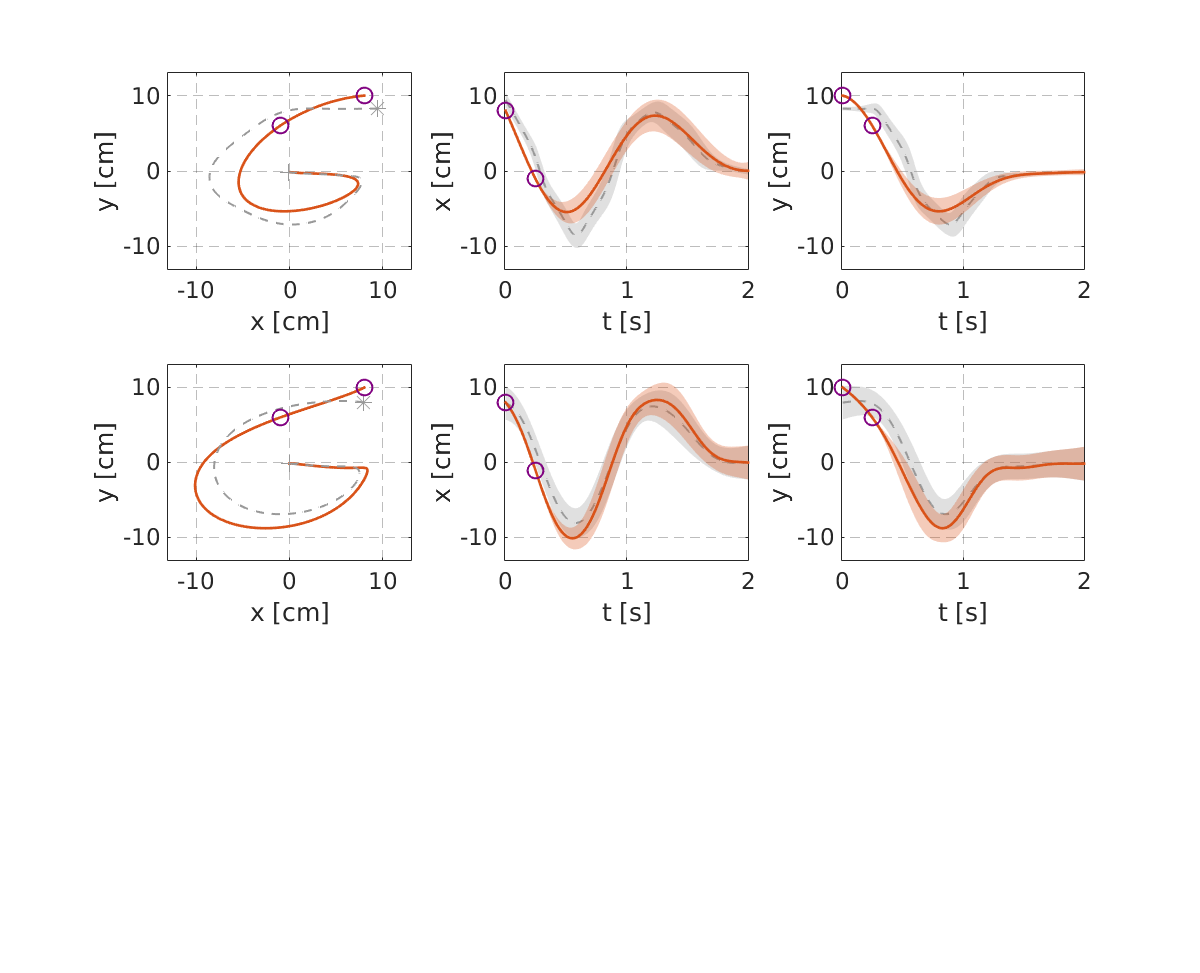}
		\put (-412.75,161) {KMP}
		\put (-416,55) {ProMP}}
	\subfigure[Trajectory modulation with one via-point and one end-point.]{ 
		\includegraphics[width=0.80\textwidth,bb=0.0cm 9.2cm 29cm 24cm,clip]{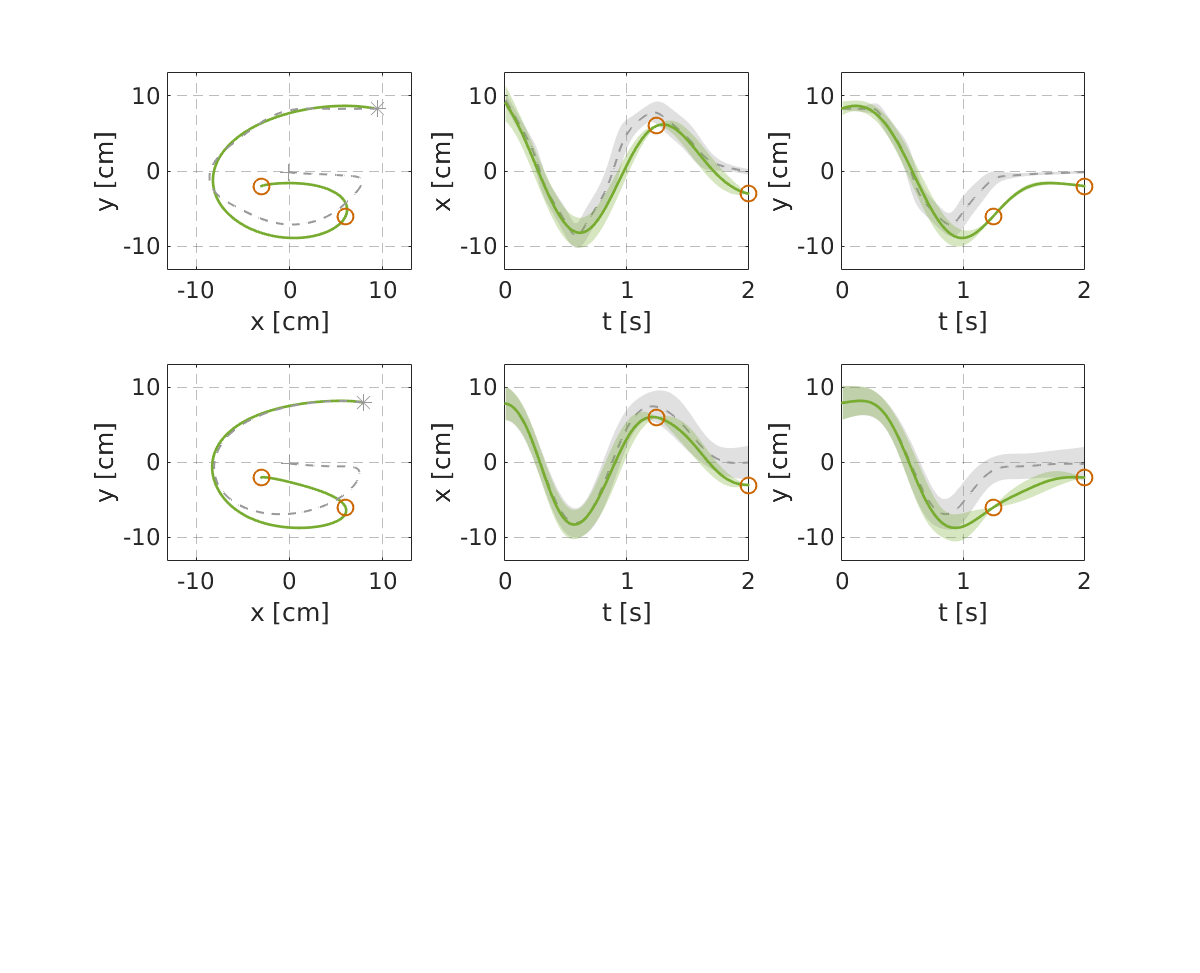}
		\put (-412.75,161) {KMP}
		\put (-416,55) {ProMP}}
	\subfigure[Superposition of two probabilistic reference trajectories.]{ 
		\includegraphics[width=0.80\textwidth,bb=0.0cm 9.2cm 29cm 24cm,clip]{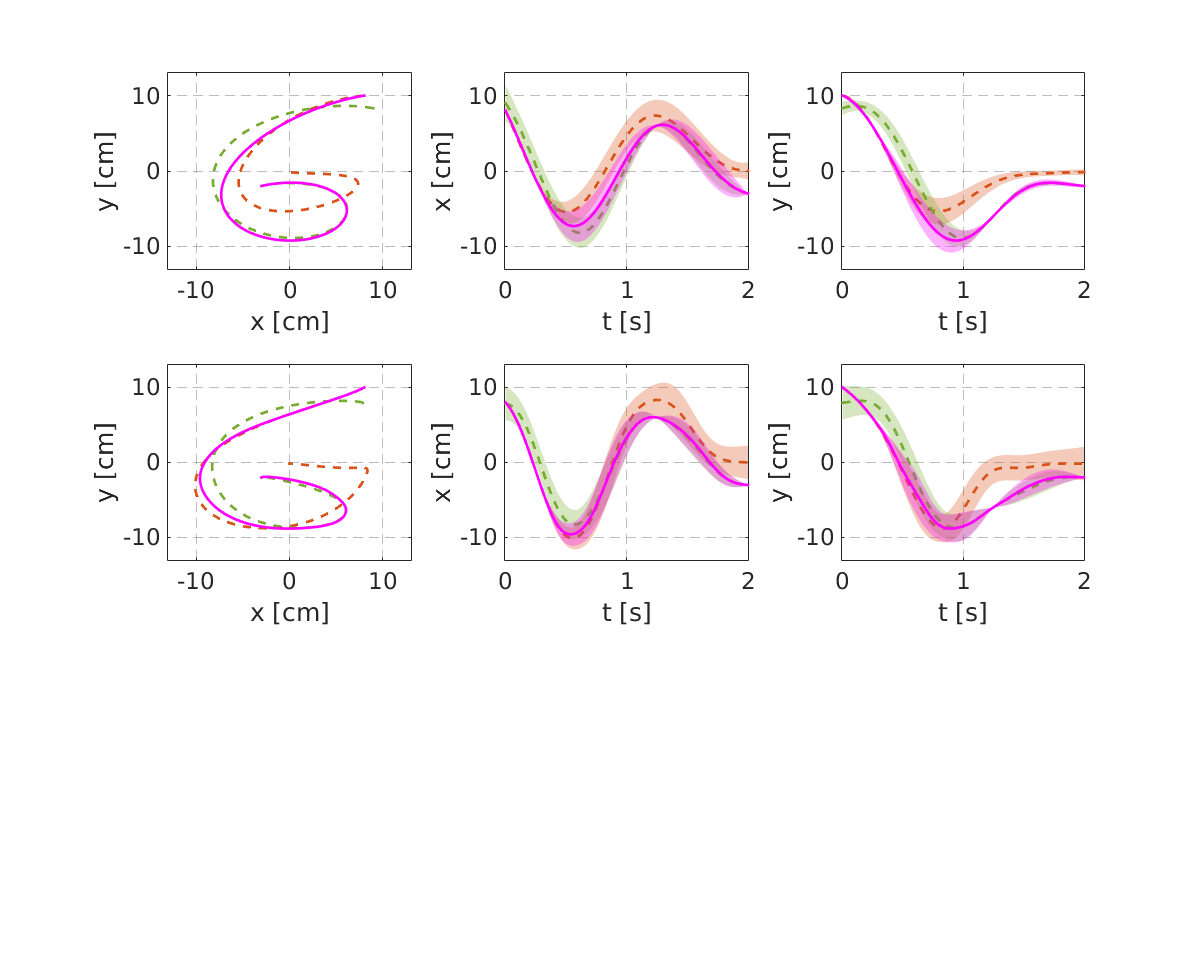}
		\put (-412.75,161) {KMP}
		\put (-416,55) {ProMP}}
	\caption{Different cases of trajectory modulation using KMP and ProMP. \emph{(a)--(b)} show trajectories (red and green curves) that are adapted to go through different desired points (depicted by circles). The gray dashed curves represent the mean of probabilistic reference trajectories for KMP and ProMP, while 
	the shaded areas depict the standard deviation.
	\emph{(c)} shows the superposition of various reference trajectories, where the dashed red and green curves correspond to the adapted trajectories in \emph{(a)} and \emph{(b)}, respectively.  
	The resulting trajectory is displayed in solid pink curve.} 
	\label{fig:viapoint:compare} 
\end{figure*} 

\begin{figure*} \centering 
	\subfigure[]{ 
		\includegraphics[width=0.31\textwidth]{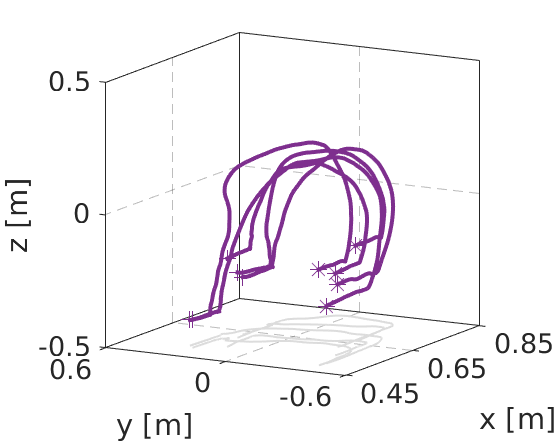}}  
	\subfigure[]{ 
		\includegraphics[width=0.31\textwidth]{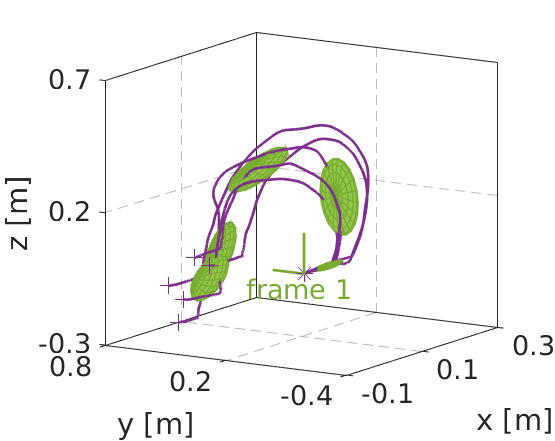}} 
	\subfigure[]{ 
		\includegraphics[width=0.31\textwidth]{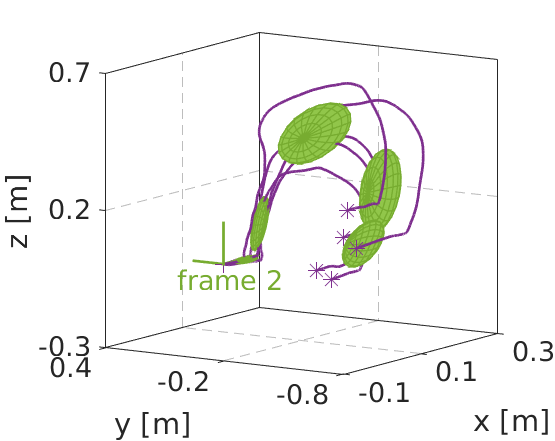}}
	\caption{Demonstrations of the transportation task as well as GMM modeling of local trajectories. (\emph{a}) shows the demonstrated trajectories (plotted by purple curves), where gray curves correspond to the projection of demonstrated trajectories into the $x$--$y$ plane. `$\ast$' and `+' denote the starting and ending points of trajectories, respectively. (\emph{b})-(\emph{c}) depict GMM modeling of local trajectories, where local trajectories are obtained by projecting demonstrations into two local frames, respectively. } 
	\label{fig:project:gmm} 
\end{figure*}

\subsection{Trajectory Modulation/Superposition}
\label{subsec:traj:modulate}

We first evaluate our approach using five  trajectories of the handwritten letter `G'\footnote{These trajectories are obtained from \cite{Calinon2017}.}, 
as shown in Figure~\ref{fig:g:demos}(\emph{a}).
These demonstrations are encoded by GMM with input $t$ and output $\vec{\xi}(t)$ being the 2-D position $[x(t)\, y(t)]^{\trsp}$. Subsequently, a probabilistic reference trajectory is retrieved through GMR, as depicted in Figure~\ref{fig:g:demos}(\emph{b})--(\emph{c}), where the position values from the reference trajectory are shown. This probabilistic reference trajectory along with the input is used to initialize KMP as described in Section~\ref{subsec:time:tmp}, which uses a Gaussian kernel ${k(t_i,t_j)=exp(-\ell (t_i-t_j)^{2})}$ with hyperparameter $\ell>0$. The relevant parameters for KMP are set as $\ell=2$ and $\lambda=1$. 

For comparison purposes, ProMP is evaluated as well, where 21 Gaussian basis functions chosen empirically are used. For each demonstration, we employ the regularized least squares method to estimate the weights $\vec{w}\in \mathbb{R}^{42}$ of the corresponding basis functions. Subsequently, the probability distribution $\mathcal{P}(\vec{\mu}_w,\vec{\Sigma}_w)$ that is computed through maximum likelihood estimation \citep{Paraschos2015} is used to initialize ProMP.  Due to the number of demonstrations being significantly lower when compared to the dimension of $\vec{w}$, a diagonal regularized term is added to $\vec{\Sigma}_w$ in order to avoid singular estimations. 
 
Figure~\ref{fig:viapoint:compare} displays different trajectory modulation cases using KMP and ProMP. We test not only cases in which new requirements arise in the form of via-points and start-points/end-points, but also the scenario of mixing different reference trajectories\footnote{We here only consider position requirements, but velocity constraints can also be directly incorporated in desired points.}. It can be observed from Figure~\ref{fig:viapoint:compare}(\emph{a})--(\emph{b}) that both KMP and ProMP successfully generate trajectories that fulfill the new requirements. For the case of trajectory superposition in Figure~\ref{fig:viapoint:compare}(\emph{c}), we consider the adapted trajectories in Figure~\ref{fig:viapoint:compare}(\emph{a}) and (\emph{b}) as candidate reference trajectories and assign them with the priorities ${\gamma_{t,1}=\exp(-t)}$ and $\gamma_{t,2}=1-\exp(-t)$, respectively. Note that $\gamma_{t,1}$ and $\gamma_{t,2}$ correspond to monotonically decreasing and increasing functions, respectively. As depicted in Figure~\ref{fig:viapoint:compare}(\emph{c}), the mixed trajectory (solid pink curve) indeed switches from the first to the second reference trajectory. 

Despite KMP and ProMP perform similarly, the key difference between them lies on the determination of basis functions. In contrast to ProMP that requires explicit basis functions, KMP is a non-parametric method that does not depend on explicit basis functions. This difference proves to be substantially crucial for tasks where the robot actions are driven by a high-dimensional input. We will show this effect in the 3rd-hand task which is associated with a 6-D input, where the implementation of ProMP becomes difficult since a large number of basis functions need to be defined.

\subsection{Extrapolation with Local-KMPs}
\label{subsec:extra:evaluate}

We evaluate the extrapolation capabilities of local-KMPs in an application with a new set of desired points (start-, via- and end-points) lying far away from the area covered by the original demonstrations, in contrast to the experiment reported in Section~\ref{subsec:traj:modulate}. Note that ProMP does not consider any task-parameterization, and therefore the extrapolation capability is limited (see \cite{Havoutis} for a discussion). Thus, we only evaluate our approach here.

We study a collaborative object transportation task, where the robot assists a human to carry an object from a starting point to an ending location. Five demonstrations
in the robot base frame are used for the training of local-KMPs (see Figure~\ref{fig:project:gmm}(\emph{a})). We consider time $t$ as the input, and the 3-D Cartesian position $[x(t)\, y(t)\, z(t)]^{\trsp}$ of the robot end-effector as the output $\vec{\xi}(t)$.
For the implementation of local-KMPs, we define two frames located at the initial and the final locations of the transportation trajectories (as depicted in Figure~\ref{fig:project:gmm}(\emph{b})--(\emph{c})), 
similarly to \cite{Leonel15}, 
which are then used to extract the local motion patterns. 

\begin{figure} \centering 	
	\includegraphics[width=0.72\columnwidth]{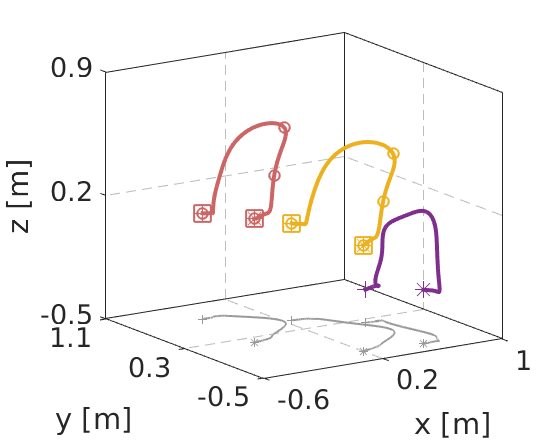}
	\caption{Extrapolation evaluations of local-KMPs for new starting and ending locations in the transportation task. The purple curve represents the mean of the original probabilistic reference trajectory for KMP, while the red and yellow trajectories show the extrapolation cases. Circles represent desired points describing additional task requirements. Squares denote desired starting and ending locations of the transportation task. Gray curves depict the projection of trajectories into the $x$--$y$ plane.} 
	\label{fig:extra:compare} 
\end{figure}

We consider two extrapolation tests, where the starting and ending locations are different from the demonstrated ones. 
In the first test, we study the transportation from ${\vec{p}_s\!=\![-0.2 \; 0.2\;0.2]^{\trsp}}$ to ${\vec{p}_e\!=\![-0.15\; 0.8\; 0.1]^{\trsp}}$. In the second test, we evaluate the extrapolation with ${\vec{p}_s\!=\![0.2\; -\!0.3 \; 0.1]^{\trsp}}$ and ${\vec{p}_e\!=\![0.25\; 0.5\; 0.05]^{\trsp}}$. Note that all locations are described with respect to the robot base frame. In addition to the desired starting and ending locations in the transportation task, we also introduce additional position constraints which require the robot passing through two via-points (plotted by circles in Figure~\ref{fig:extra:compare}).
The extrapolation of local-KMPs for these new situations is achieved according to Algorithm \ref{algorithm:local-kmp}, where the Gaussian kernel is used. For each test, the local frames are set as ${\vec{A}^{(1)}=\vec{A}^{(2)}=\vec{I}_3}$, $\vec{b}^{(1)}=\vec{p}_s$ and $\vec{b}^{(2)}=\vec{p}_e$. The related KMP parameters are $\ell=0.5$ and $\lambda=10$.
Figure~\ref{fig:extra:compare} shows that local-KMPs successfully extrapolate to new frame locations and lead the robot to go through various new desired points while maintaining the shape of the demonstrated trajectories. 

Note that the environment might drastically change from demonstrations to final execution, so the capability of modulating the demonstrated trajectories to go through new points is important in many applications. In this sense, local-KMPs prove superior to other local-frame approaches such as those exploited in \cite{Leonel15, Calinon2016}, which do not consider trajectory modulation.

\subsection{Force-based Trajectory Adaptation}
\label{subsec:force:ada}
Through kinesthetic teaching, humans are able to provide the robot with initial feasible trajectories. However, this procedure does not account for unpredicted situations. For instance, when the robot is moving towards a target, undesired circumstances such as obstacles occupying the robot workspace might appear, which requires the robot to avoid possible collisions. 
Since humans have reliable reactions over dynamic environments, we here propose to use the human supervision to adapt the robot trajectory when the environment changes. In particular, we use a force sensor installed at the end-effector of the robot in order to measure corrective forces exerted by the human. 

We treat the force-based adaptation problem under the KMP framework by defining new via-points as a function of the sensed forces.
Whenever the robot is about to collide with the obstacle, the user interacts physically with the end-effector and applies a corrective force. This force is used to determine a desired via-point which the robot needs to pass through in order to avoid the obstacle.
By updating the reference database using this obtained via-point through (\ref{equ:kmp:update}), KMP can generate an adapted trajectory that fulfills the via-point constraint.

For the human interaction at time $t$, given the robot Cartesian position $\vec{p}_t$ and the sensed force $\vec{F}_t$, the first desired datapoint is defined as:
${\bar{t}_1=t+\delta_t}$ and ${\bar{\vec{p}}_{1}=\vec{p}_{t}+\vec{K}_f \vec{F}_t}$, where $\delta_t>0$ controls the regulation time and $\vec{K}_f>0$ determines the adaptation proportion for the robot trajectory. In order to avoid undesired trajectory modulations caused by the force sensor noise, we introduce a force threshold $F_{thre}$ and add the new force-based via-point to the reference trajectory only when $||\vec{F}_t||>F_{thre}$.
Note that the adapted trajectory might be far away from the previous planned trajectory due to the new via-point, we hence consider adding $\vec{p}_t$ as the second desired point so as to ensure a smooth trajectory for the robot. Doing so, for each interaction, we define the second desired point as $\bar{t}_2=t$ and $\bar{\vec{p}}_{2}=\vec{p}_t$.

\begin{figure} \centering
	\includegraphics[width=0.92\columnwidth]{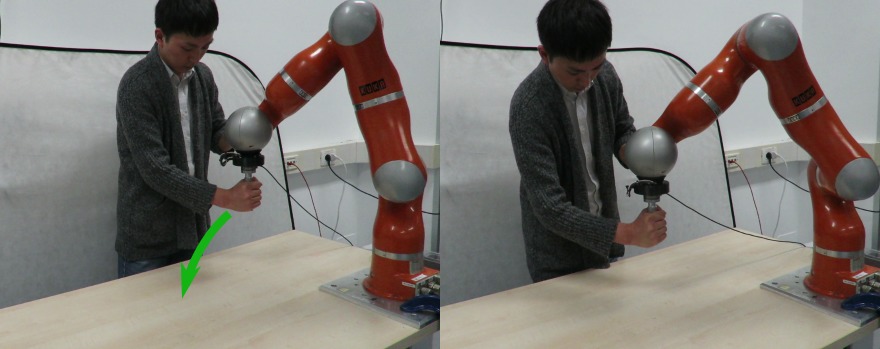}
	\caption{Kinesthetic teaching of the reaching task on the KUKA robot, where demonstrations comprising time and end-effector Cartesian position are collected. The green arrow shows the motion direction of the robot.} 
	\label{fig:force:humanDemo} 
\end{figure} 

\begin{figure} \centering 
	\includegraphics[width=0.90\columnwidth]{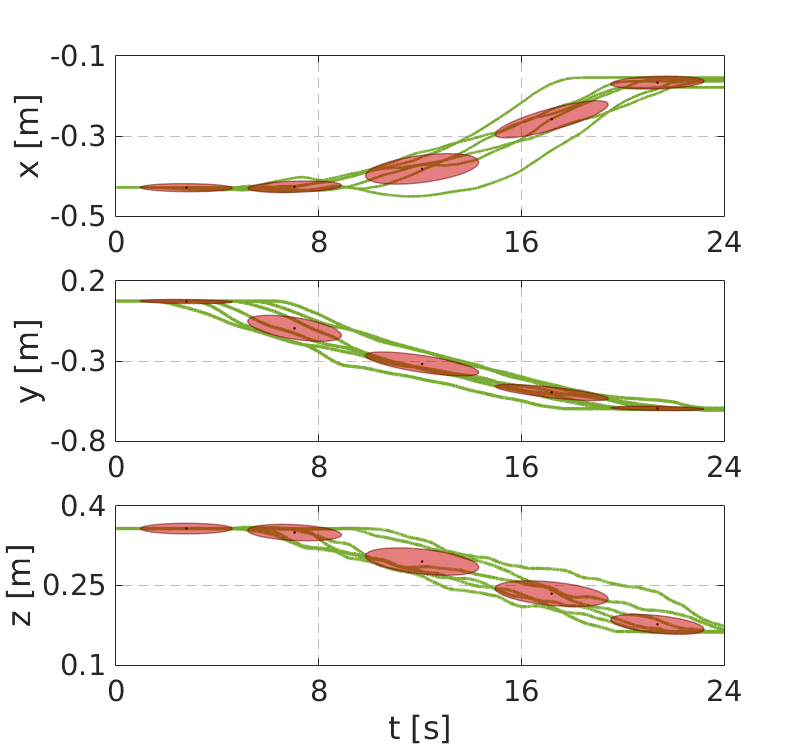}
	\caption{GMM modeling of demonstrations for the force-based adaptation task, where the green curves represent demonstrated trajectories and ellipses depict Gaussian components.} 
	\label{fig:force:demos} 
\end{figure}

\begin{figure*} \centering
	\includegraphics[width=2.03\columnwidth]{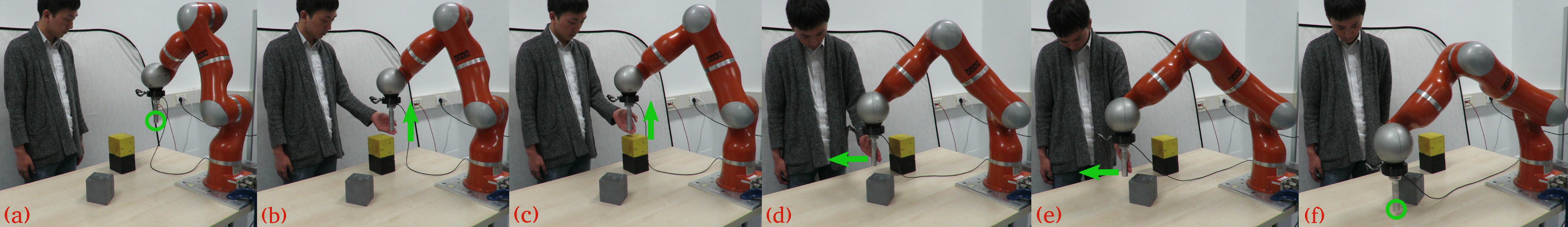}
	\caption{Snapshots of the force-based trajectory adaptations, where the force exerted by the human is used to determine the via-points for the robot, which ensures collision avoidance. (a) and (f) correspond to the initial and final states of the robot, where circles depict the initial and final positions, respectively. Figures (b)--(e) show human interactions with the green arrows depicting the directions of corrective force. }
	\label{fig:force:robot} 
\end{figure*}

\begin{figure*} \centering 
	\includegraphics[width=1.9\columnwidth,bb=4.0cm 0cm 45cm 21.5cm,clip]{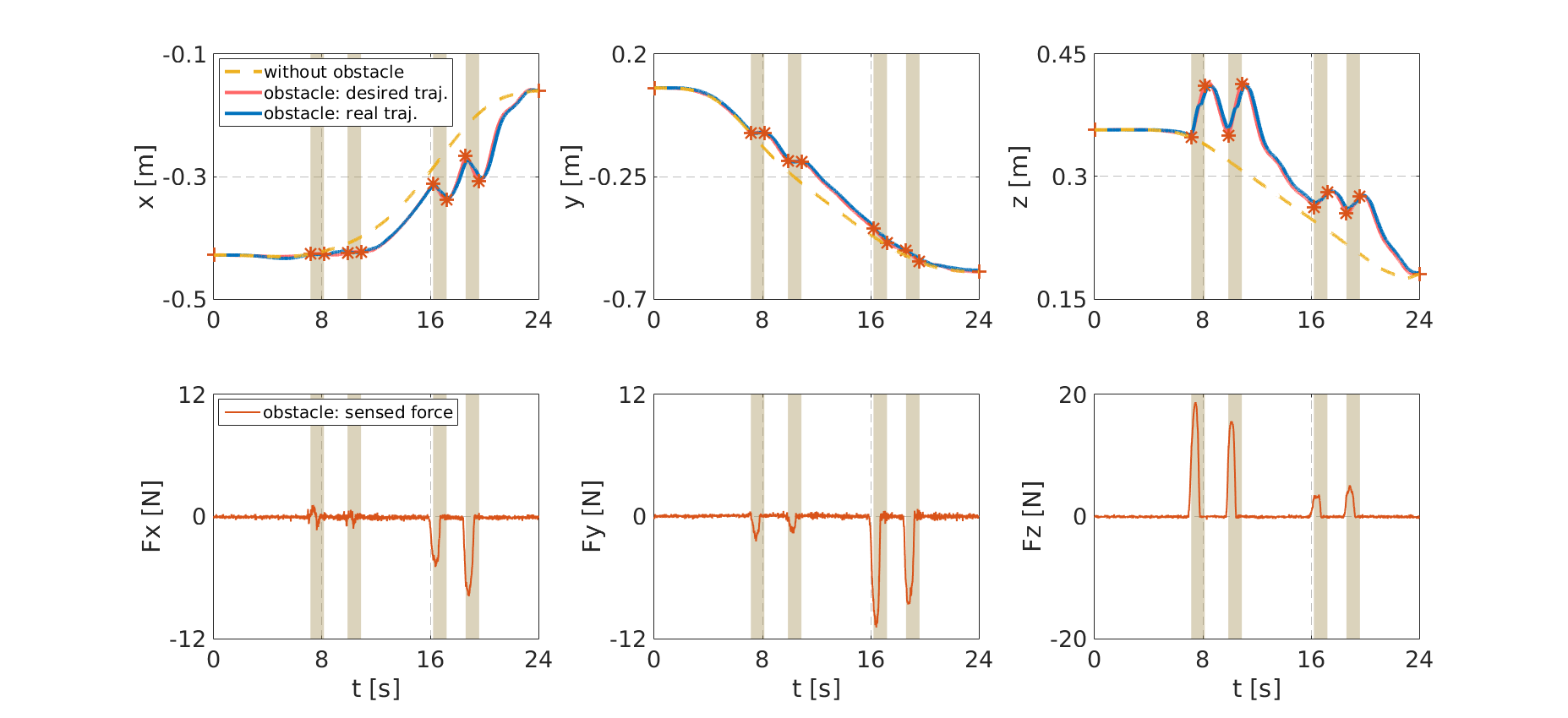}
	\caption{\emph{Top row}: the desired trajectory (generated by KMP) and the real robot trajectory, where `$\ast$' represents the force-based desired points and `+' corresponds to the initial and final locations for the robot. For comparison, we also provide the desired trajectory predicted by KMP without obstacles (i.e., without human intervention). The shaded areas show the regulation durations for various human interventions. \emph{Bottom row}: the force measured at the end-effector of the KUKA robot.} 
	\label{fig:force:adaTraj} 
\end{figure*}

In order to evaluate the adaptation capability of KMP, we consider a reaching task where unpredicted obstacles will intersect the robot movement path. First, we collect six demonstrations (as depicted in Figure~\ref{fig:force:humanDemo}) comprising time input $t$ and output $\vec{\xi}(t)$ being the 3-D Cartesian position ${[x(t)\, y(t)\, z(t)]^{\trsp}}$. Note that obstacles are not placed in the training phase. The collected data is fitted using GMM (plotted in Figure~\ref{fig:force:demos}) so as to retrieve a reference database, which is subsequently used to initialize KMP. 
Then, during the evaluation phase,
two obstacles whose locations intersect the robot path are placed on the table, as shown in Figure~\ref{fig:force:robot}.
In addition to the via-points that will be added through physical interaction,
we add the initial and target locations for the robot as desired points beforehand, where the initial location corresponds to the robot position before starting moving.
The relevant parameters are $\vec{K}_f\!\!=\!\!0.006\vec{I}_{3}$, $\delta_t=1s$, $F_{thre}=10N$, $\ell=0.15$ and ${\lambda=0.3}$.

The trajectory that is generated by KMP according to various desired points
as well as the real robot trajectory are depicted in Figure~\ref{fig:force:adaTraj}.
We can observe that for each obstacle the robot trajectory is adapted twice. In the first two adaptations (around $8s$ and $11s$), the corrective force is dominant along the $z$ direction, while in the last two adaptations (around $17s$ and $20s$), the force has a larger component along the $x$ and $y$ directions. For all cases, KMP successfully adapts the end-effector trajectory according to the measured forces. 

Note that, even without human interaction, the proposed scheme can also help the robot replan its trajectory when it touches the obstacles, where the collision force takes the role of the human correction and guides the robot to move away from the obstacles. 
Thus, with KMP the robot is capable of autonomously adapting its trajectory through low-impact collisions, whose tolerated force can be regulated using $F_{thre}$.
Supplementary material includes a video of experiments using the human corrective force and the obstacle collision force.

\begin{figure*} \centering
	\includegraphics[width=2.03\columnwidth]{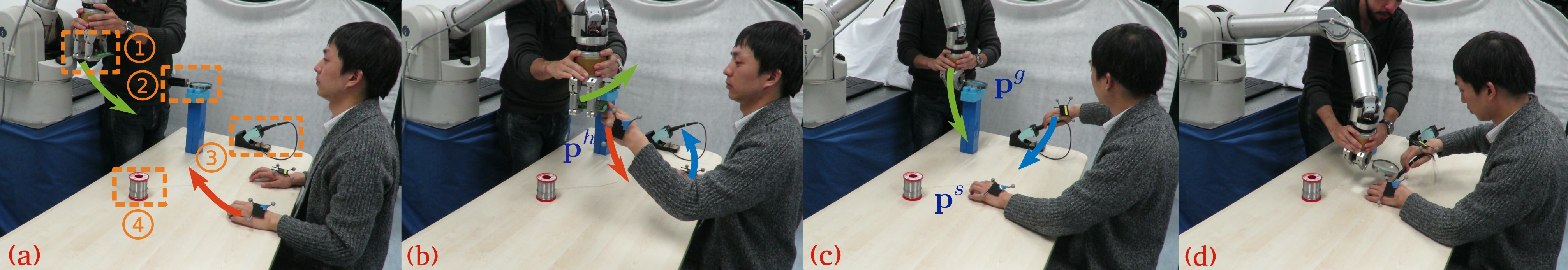}
	\caption{The 3rd-hand task in the soldering environment with the Barrett WAM robot. \emph{(a)} shows the initial states of the user hands and the robot end-effector (the 3rd hand in this experiment). \textcircled{1}--\textcircled{4} separately correspond to the circuit board (held by the robot), magnifying glass, soldering iron and solder. \emph{(b)} corresponds to the handover of the circuit board. \emph{(c)} shows the robot grasping of the magnifying glass. 
		\emph{(d)} depicts the final scenario of the soldering task using both of the user hands and the robot end-effector. Red, blue and green arrows depict the movement directions of the user left hand, right hand and the robot end-effector, respectively. }
	\label{fig:3rHand:demos:robot} 
\end{figure*}

\subsection{3rd-hand Task}
\label{subsec:3rd:hand}

So far the reported experiments have shown the performances of KMP by learning various time-driven trajectories. We now consider a different task which requires a 6-D input, in particular a robot-assisted soldering scenario.
As shown in Figure~\ref{fig:3rHand:demos:robot}, the task proceeds as follows: \emph{(1)} the robot needs to hand over a circuit board to the user at the \emph{handover location} $\vec{p}^{h}$ (Figure~\ref{fig:3rHand:demos:robot}\emph{(b)}), where the left hand of the user is used. \emph{(2)} the user moves his left hand to place the circuit board at the \emph{soldering location} $\vec{p}^{s}$ and simultaneously moves his right hand towards the soldering iron and then grasps it. Meanwhile, the robot is required to move towards the magnifying glass and grasp it at the \emph{magnifying glass location} $\vec{p}^{g}$ (Figure~\ref{fig:3rHand:demos:robot}\emph{(c)}). \emph{(3)} the user moves his right hand to the soldering location so as to repair the circuit board. Meanwhile, the robot, holding the magnifying glass, moves towards the soldering place in order to allow the user to take a better look at the small components of the board (Figure~\ref{fig:3rHand:demos:robot}\emph{(d)}).

Let us denote $\vec{p}^{\mathcal{H}_l}$, $\vec{p}^{\mathcal{H}_r}$ and $\vec{p}^{\mathcal{R}}$ as positions of the user left hand, right hand and robot end-effector (i.e., the ``third hand''), respectively. 
Since the robot is required to react properly according to the user hand positions, we formulate the 3rd-hand task as the prediction of the robot end-effector position according to the user hand positions. In other words, in the prediction problem we consider ${\vec{s}=\{\vec{p}^{\mathcal{H}_l}, \vec{p}^{\mathcal{H}_r}\}}$ as the input (6-D) and $\vec{\xi}(\vec{s})=\vec{p}^{\mathcal{R}}$ as the output (3-D) .  

\begin{figure} \centering
	\subfigure[]{  
		\includegraphics[width=0.32\textwidth]{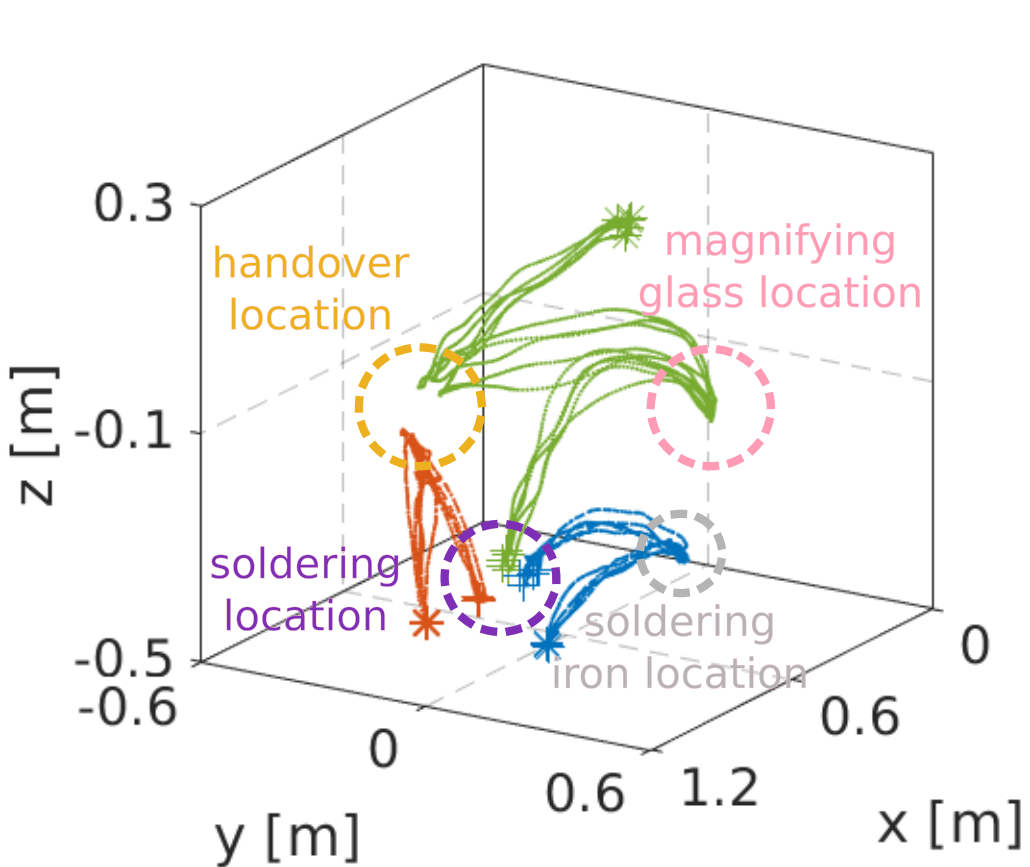}}			
	\caption{
		Demonstrations for the 3rd-hand task,
		where the red and blue curves respectively correspond to the user left and right hands, while the green curves represent the demonstrated trajectories for the robot. The `$\ast$' and `+' mark the starting and ending points of various trajectories, respectively.} 
	\label{fig:3rdHand:task:demo} 
\end{figure}

Following the procedure illustrated in Figure~\ref{fig:3rHand:demos:robot}, we collect five demonstrations comprising $\{\vec{p}^{\mathcal{H}_l}\!,\vec{p}^{\mathcal{H}_r}\!,\vec{p}^{\mathcal{R}}\!\}$ for training KMP, as shown in Figure~\ref{fig:3rdHand:task:demo}.  
Note that the teacher only gets involved in the training phase.
We fit the collected data using GMM, and subsequently extract a probabilistic reference trajectory using GMR,
where the input for the probabilistic reference trajectory is sampled from the marginal probability distribution $\mathcal{P}(\vec{s})$, since in this scenario the exact input is unknown (unlike time $t$ in previous experiments). The Gaussian kernel 
is also employed in KMP, whose hyperparameters are set to $\ell=0.5$ and $\lambda=2$.

\begin{figure*} \centering
	\includegraphics[width=2.03\columnwidth]{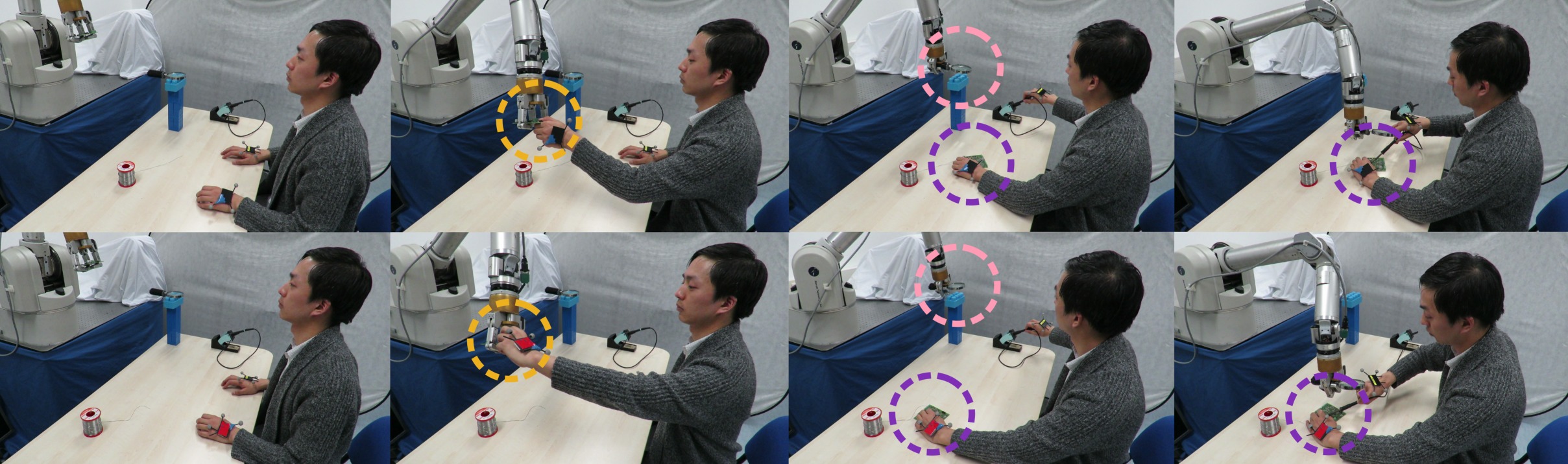}	
	\caption{Snapshots of reproduction and adaptation using KMP. \emph{Top row} shows the reproduction case using the learned reference database without adaptation. \emph{Bottom row} displays the adaptation case using the new reference database which is updated using three new desired points: new handover, magnifying glass and soldering locations depicted as dashed circles (notice the difference with respect to the top row).} 
	\label{fig:3rdHand:eva} 
\end{figure*}

\begin{figure} \centering
	\subfigure[]{ 
		\includegraphics[width=0.32\textwidth]{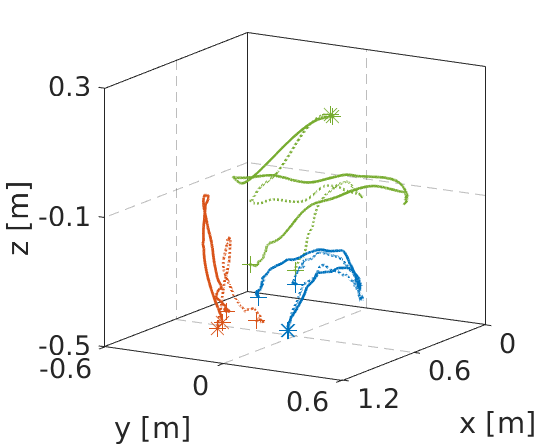}}			
	\caption{
		The reproduction (dotted curves) and adaptation (solid curves) capabilities of KMP in the 3rd-hand task, where the user left-hand and right-hand trajectories (red and blue curves) are used to retrieve the robot end-effector trajectory (green curves).} 
	\label{fig:3rdHand:task:eva} 
\end{figure}

Two evaluations are carried out to evaluate KMP in this scenario. 
First, we employ the learned reference database without adaptation so as to verify the reproduction ability of KMP, as shown in Figure~\ref{fig:3rdHand:eva} (\emph{top row}). 
The user left- and right-hand trajectories as well as the real robot trajectory, depicted in Figure~\ref{fig:3rdHand:eva} (\emph{top row}), are plotted in Figure~\ref{fig:3rdHand:task:eva} (dotted curves), where the desired trajectory for robot end-effector is generated by KMP. We can observe that KMP maintains the shape of the demonstrated trajectories for the robot while accomplishing the soldering task. Second, we evaluate the adaptation capability of KMP by adjusting the handover location 
$\vec{p}^{h}$, the magnifying glass location $\vec{p}^{g}$ as well as the soldering location $\vec{p}^{s}$, as illustrated in Figure~\ref{fig:3rdHand:eva} (\emph{bottom row}). 
Note that these new locations are unseen in the demonstrations, thus we consider them as new via-point/end-point constraints within the KMP framework. 

To take the handover as an example, we can define a via-point (associated with input) as
$\{\bar{\vec{p}}^{\mathcal{H}_l}_{1},\bar{\vec{p}}^{\mathcal{H}_r}_{1},\bar{\vec{p}}^{\mathcal{R}}_{1}\}$, where
$\bar{\vec{p}}^{\mathcal{H}_l}_{1}=\vec{p}^{h}$, $\bar{\vec{p}}^{\mathcal{H}_r}_{1}=\vec{p}^{\mathcal{H}_r}_{ini}$ and $\bar{\vec{p}}^{\mathcal{R}}_{1}=\vec{p}^{h}$, which implies that the robot should reach the new handover location $\vec{p}^{h}$ when the user left hand arrives at $\vec{p}^{h}$ and the user right hand stays at its initial position $\vec{p}^{\mathcal{H}_r}_{ini}$.
Similarly, we can define additional via- and end-points to ensure that the robot grasps the magnifying glass at a new location $\vec{p}^{g}$ and assists the user at a new location $\vec{p}^{s}$. Thus, two via-points and one end-point are used to update the original reference database according to (\ref{equ:kmp:update}) so as to address the three adaptation situations.
Figure~\ref{fig:3rdHand:task:eva} shows the adaptations of the robot trajectory (green solid curve) in accordance with the user hand trajectories (red and blue solid curves).
It can be seen that the robot trajectory is indeed modulated towards the new handover, magnifying glass and soldering locations, showing the capability of KMP to adapt trajectories associated with high-dimensional inputs.

It is worth pointing out that the entire soldering task is accomplished by a single KMP without any trajectory segmentation for different subtasks, thus allowing for a straightforward learning of several sequential subtasks. Moreover, KMP makes the adaptation of learned skills associated with high-dimensional inputs feasible.  Also, KMP is driven by the user hand positions, which allows for slower/faster hand movements since the prediction of KMP does not depend on time, hence alleviating the typical problem of time-alignment in human-robot collaborations. For details on the 3rd-hand experiments, please refer to the video in the supplementary material.

\section{Related Work}
\label{sec:relative:work}

In light of the reliable temporal and spatial generalization, DMP \citep{Ijspeert} has achieved remarkable success in a vast range of applications.
In addition, many variants of DMP have been developed for specific circumstances, such as stylistic DMP \citep{Matsubara}, task-parameterized DMP \citep{Pervez} and combined DMP \citep{Pastor}. However, due to the spring-damper dynamics, DMP converges to the target position with zero velocity, which prevents its application to cases with velocity requirements (e.g., the striking/batting movement). Besides, DMP does not provide a straightforward way to incorporate desired via-points. 

By exploiting the properties of Gaussian distributions, ProMP \citep{Paraschos} allows for trajectory adaptations with via-points and end-points simultaneously. The similarities between DMP and ProMP lie on the fact that both methods need the explicit definition of basis functions and are aimed at learning time-driven trajectories. As a consequence, when we encounter trajectories with  high-dimensional inputs (e.g., human hand position and posture in human-robot collaboration scenarios), the selection of basis functions in DMP and ProMP becomes difficult and thus undesired. 

In contrast to DMP and ProMP, GMM/GMR based learning algorithms \citep{Muhlig,Calinon2007} have been proven effective in encoding demonstrations with high-dimensional inputs. However, the large number of variables arising in GMM makes the re-optimization of GMM expensive, which therefore prevents its use in unstructured environments where robot adaptation capabilities are imperative. 

KMP provides several advantages compared to the aforementioned works. Unlike GMM/GMR, KMP is capable of adapting trajectories towards various via-points/end-points without the optimization of high-dimensional hyperparameters. Unlike DMP and ProMP, KMP alleviates the need of explicit basis functions due to its kernel treatment, and thus can be easily implemented for problems with high-dimensional inputs and outputs. 

It is noted that the training of DMP only needs a single demonstration, while ProMP, GMM and KMP require a set of trajectories. In contrast to the learning of a single demonstration, the exploitation of multiple demonstrations makes the extraction of probabilistic properties of human skills possible. In this context, demonstrations have been exploited using the covariance-weighted strategy, as in trajectory-GMM \citep{Calinon2016}, linear quadratic regulators (LQR) \citep{Leonel15},
movement similarity criterion \citep{Muhlig} and demonstration-guided trajectory optimization \citep{Osa}. Note that the mean minimization subproblem as formulated in (\ref{equ:kl:mean:cost}) also uses the covariance to weigh the cost, sharing the same spirit of the aforementioned results.

Similarly to our approach, information theory has also been exploited in different robot learning techniques. As an effective way to measure the distance between two probabilistic distributions, KL-divergence was exploited in policy search \citep{Peters, Kahn}, trajectory optimization \citep{Levine} and imitation learning \citep{Englert}.
In \cite{Englert} KL-divergence was used to measure the difference between the distributions of demonstrations and the predicted robot trajectories (obtained from a control policy and a Gaussian process forward model), and subsequently the probabilistic inference for learning control \citep{Deisenroth} was employed to iteratively minimize the KL-divergence so as to find the optimal policy parameters. It is noted that this KL-divergence formulation makes the derivations of analytical solution intractable. In this article, we formulate the trajectory matching problem as (\ref{equ:kl:cost:ini:temp}), which allows us to separate the mean and covariance subproblems and derive closed-form solutions for them separately.

\section{Discussion} 
\label{sec:discuss}
While both KMP and ProMP \citep{Paraschos} learn the probabilistic properties of demonstrations, we here discuss their similarities and possible shortcomings in detail.  
For the KMP, 
imitation learning is formulated as an optimization problem (Section \ref{subsubsec:kl}), where the optimal distribution $\mathcal{N}({\vec{\mu}}_w^{*},{\vec{\Sigma}}_w^{*})$ of $\vec{w}$ is derived by minimizing the information-loss between the parametric trajectory and the demonstrations. Specifically, the mean minimization subproblem (\ref{equ:kl:mean:cost}) can be viewed as the problem of maximizing the posterior $\prod_{n=1}^{N} \mathcal{P}(\vec{\Theta}(\vec{s}_n)^{\trsp} \vec{\mu}_w|\hat{\vec{\mu}}_n,\hat{\vec{\Sigma}}_n)$. 
In contrast, ProMP formulates the problem of imitation learning as an estimation of the probability distribution of movement pattern $\vec{w}$ (i.e., $\vec{w} \sim \mathcal{N}(\vec{\mu}_w,\vec{\Sigma}_w)$), which is essentially equivalent to the maximization of the likelihood $\prod_{h=1}^{H} \prod_{n=1}^{N} \mathcal{P}({\vec{\xi}}_{n,h}|\vec{\Theta}(\vec{s}_n)^{\trsp}\vec{\mu}_w,\vec{\Theta}(\vec{s}_n)^{\trsp}\vec{\Sigma}_w \vec{\Theta}(\vec{s}_n))$. 
To solve this maximization problem, the regularized least-squares is first used for each demonstration so as to estimate its corresponding movement pattern vector \citep{Paraschos2015}, where basis functions are used to fit these demonstrations. Subsequently, using the movement patterns extracted from demonstrations, the distribution $\mathcal{P}(\vec{w})$ is determined by using the maximum likelihood estimation.

A direct problem in ProMP is the estimation of $\mathcal{P}(\vec{w})$.
If the dimension of $\vec{w}$ (i.e., $B\mathcal{O}$) is too high compared to the number of demonstrations $H$, a singular covariance $\vec{\Sigma}_w$ might appear. For this reason, learning movements with ProMP typically requires a high number of demonstrations. In contrast, KMP needs a probabilistic reference trajectory, which is derived from the joint probability distribution of $\{\vec{s},\vec{\xi}\}$ that is typically characterized by a lower dimensionality (i.e., $\mathcal{I}+\mathcal{O}$). 
Another problem in ProMP comes up with demonstrations with high dimensional input $\vec{s}$, where the number of basis functions increases often exponentially, which is the typical curse of dimensionality (see also the discussion on the disadvantages of fixed basis functions in \cite{Bishop}). In contrast, KMP is combined with a kernel function, alleviating the need for basis functions, while inheriting all the potentials and expressiveness of kernel-based methods.

There are several possible extensions for KMP. First, similarly to most regression algorithms, the computation complexity of KMP increases with the size of training data (i.e., the reference database in our case). One possible solution could be the use of partial training data so as to build a sparse model \citep{Bishop}.
Second, even though we have shown the capability of KMP on trajectory adaptation, the choice of desired points is rather empirical. For more complicated situations where we have no (or minor) prior information, the search of optimal desired points could be useful. To address this problem, RL algorithms could be employed to find appropriate new via-points that fulfill the relevant task requirements that can be encapsulated by cost functions. Third, since KMP predicts mean and covariance of the trajectory simultaneously, it may be exploited in approaches that combine optimal control and probabilistic learning methods \citep{Medina}. For example, the mean and covariance can be respectively used as the desired trajectory and the weighted matrix for tracking errors in LQR \citep{Calinon2016}.
Finally, besides the frequently used Gaussian kernel, the exploitation of various kernels \citep{Hofmann} could be promising in the future research.

\section{Conclusions} 
\label{sec:conclusion}
We have proposed a novel formulation of robot movement primitives that incorporates a kernel-based treatment into the process of minimizing the information-loss in imitation learning. Our approach KMP is capable of preserving the probabilistic properties of human demonstrations, adapting trajectories to different unseen situations described by new temporal or spatial requirements and mixing different trajectories. The proposed method was extended to deal with local frames, which provides the robot with reliable extrapolation capabilities. Since KMP is essentially a kernel-based non-parametric approach, it overcomes several limitations of state-of-the-art methods, being able to model complex and high dimensional trajectories. Through extensive evaluations in simulations and real robotic systems, we showed that KMP performs well in a wide range of applications such as time-driven movements and human-robot collaboration scenarios.

\section*{Acknowledgments}
We thank Fares J. Abu-Dakka, Luka Peternel and Martijn J. A. Zeestraten for their help on real robot experiments.

\begin{appendices}

\section{Gaussian Mixture Regression (GMR)}
\label{app:gmr}
Let us write the joint probability distribution $\mathcal{P}(\vec{s},\vec{\xi})$ as
$\left[\begin{matrix}
\vec{s}\\\vec{\xi}
\end{matrix}\right] \sim \sum_{c=1}^{C} \pi_c \mathcal{N}(\vec{\mu}_c,\vec{\Sigma}_c)$,
and decompose the mean $\vec{\mu}_c$ and covariance $\vec{\Sigma}_c$ of the $c$-th Gaussian component as
\begin{equation}
\vec{\mu}_{c}=\left[\begin{matrix}
\vec{\mu}_{c}^{s} \\ \vec{\mu}_{c}^{\xi}
\end{matrix}\right]\quad \mathrm{and} \quad \vec{\Sigma}_{c}=\left[\begin{matrix}
\vec{\Sigma}_{c}^{ss} &\vec{\Sigma}_{c}^{s \xi}\\ \vec{\Sigma}_{c}^{\xi s}   &\vec{\Sigma}_{c}^{\xi \xi}
\end{matrix}\right],
\end{equation}
where the superscripts $s$ and $\xi$ correspond to the input and output variables, respectively.

For a query input $\vec{s}_n$, the mean of its corresponding output is computed by
\begin{equation}
\hat{\vec{\mu}}_n=\mathbb{E}(\hat{\vec{\xi}}_n|\vec{s}_n)=\sum_{c=1}^{C}h_c(\vec{s}_n) \vec{\mu}_c(\vec{s}_n)
\label{equ:gmr:mean}
\end{equation}
with $h_c(\vec{s}_n)=\frac{\pi_c \mathcal{N}(\vec{s}_n|\vec{u}_{c}^{s},\vec{\Sigma}_{c}^{ss})}
{\sum_{k=1}^{C} \pi_k \mathcal{N}(\vec{s}_n|\vec{u}_{k}^{s},\vec{\Sigma}_{k}^{ss})}$
and
${\vec{\mu}_c(\vec{s}_n)=
\vec{\mu}_{c}^{\xi}+\vec{\Sigma}_{c}^{\xi s} (\vec{\Sigma}_{c}^{ss})^{-1}(\vec{s}_n-\vec{\mu}_{c}^{s})}$.
The corresponding conditional covariance is
\begin{equation}
\hat{\vec{\Sigma}}_n=\mathbb{D}(\hat{\vec{\xi}}_n|\vec{s}_n)=
\mathbb{E}(\hat{\vec{\xi}}_n \hat{\vec{\xi}}_n^{\trsp}|\vec{s}_n)-\mathbb{E}(\hat{\vec{\xi}}_n |\vec{s}_n)\mathbb{E}^{\trsp}(\hat{\vec{\xi}}_n|\vec{s}_n).
\label{equ:gmr:var}
\end{equation}
with 
$\mathbb{E}(\hat{\vec{\xi}}_n \hat{\vec{\xi}}_n^{\trsp}|\vec{s}_n)\!\!=\!\!\sum_{c=1}^{C}\!h_c(\vec{s}_n) \left(\vec{\Sigma}_c\!+\!\vec{\mu}_c(\vec{s}_n)\vec{\mu}_c(\vec{s}_n)^{\trsp}  \right)\!
$
and $\vec{\Sigma}_c=
\vec{\Sigma}_{c}^{\xi\xi}-\vec{\Sigma}_{c}^{\xi s} (\vec{\Sigma}_{c}^{ss})^{-1}\vec{\Sigma}_{c}^{s \xi}$. Thus, the conditional distribution $\hat{\vec{\xi}}_n|\vec{s}_n \sim \mathcal{N}(\hat{\vec{\mu}}_n,\hat{\vec{\Sigma}}_n)$ is determined.

\section{Interpretation of Mean Minimization Problem}
\label{app:mean:dual}
On the basis of the definition of multivariate Gaussian distribution, we have
\begin{equation}
\begin{aligned}
&\prod_{n=1}^{N}\mathcal{P}(\vec{\Theta}(\vec{s}_n)^{\trsp} \vec{\mu}_w|\hat{\vec{\mu}}_n,\hat{\vec{\Sigma}}_n)=\prod_{n=1}^{N}\frac{1}{(2\pi)^{\mathcal{O}/2}|\hat{\vec{\Sigma}}_n|^{1/2}} \\ &exp\left\{-\frac{1}{2}(\vec{\Theta}(\vec{s}_n)^{\trsp} \vec{\mu}_w-\hat{\vec{\mu}}_n)^{\trsp}\hat{\vec{\Sigma}}_n^{-1}(\vec{\Theta}(\vec{s}_n)^{\trsp} \vec{\mu}_w-\hat{\vec{\mu}}_n)\right\},
\end{aligned}
\label{equ:product:multivariate:gaussian}
\end{equation}
which can be further simplified using the logarithmic transformation, yielding
\begin{equation}
\begin{aligned}
\sum_{n=1}^{N}\!\log\mathcal{P}(\vec{\Theta}(\vec{s}_n)^{\trsp}\! \vec{\mu}_w|\hat{\vec{\mu}}_n,\hat{\vec{\Sigma}}_n)\!=\! \sum_{n=1}^{N} \!\! -\!\frac{1}{2}(\vec{\Theta}(\vec{s}_n)^{\trsp} \!\vec{\mu}_w \!-\! \hat{\vec{\mu}}_n)^{\trsp} \\ \hat{\vec{\Sigma}}_n^{-1}(\vec{\Theta}(\vec{s}_n)^{\trsp} \vec{\mu}_w-\hat{\vec{\mu}}_n)+constant.
\end{aligned}
\label{equ:logsum:multivariate:gaussian}
\end{equation}
Thus, the mean minimization problem described by (\ref{equ:kl:mean:cost}) can be interpreted as the maximization of the posterior defined in (\ref{equ:product:multivariate:gaussian}),

\section{Proof of Weighted Mean Minimization Subproblem}
\label{app:compose:mean}
The derivative of (\ref{equ:kl:mean:cost:mix}) with respect to $\vec{\mu}_w$ can be computed as
\begin{equation}
\begin{aligned}
\frac{\partial J^{S}_{ini}(\vec{\mu}_w)}{\partial \vec{\mu}_w}=
\sum_{n=1}^{N} \sum_{l=1}^{L} 2 \biggl\{\vec{\Theta}(\vec{s}_{n})\left( \frac{\hat{\vec{\Sigma}}_{n,l}}{\gamma_{n,l}} \right)^{-1} \vec{\Theta}^{\trsp}(\vec{s}_{n})\vec{\mu}_w \\
-\vec{\Theta}(\vec{s}_{n})\left( \frac{\hat{\vec{\Sigma}}_{n,l}}{\gamma_{n,l}} \right)^{-1} \hat{\vec{\mu}}_{n,l}\biggr\} \\
=\sum_{n=1}^{N} 2\vec{\Theta}(\vec{s}_{n})
\biggl\{\sum_{l=1}^{L} 
\left( \frac{\hat{\vec{\Sigma}}_{n,l}}{\gamma_{n,l}} \right)^{-1}  \biggr\}
\vec{\Theta}^{\trsp}(\vec{s}_{n})\vec{\mu}_w \\
-\sum_{n=1}^{N} 2\vec{\Theta}(\vec{s}_{n})\biggl\{\sum_{l=1}^{L} 
\left( \frac{\hat{\vec{\Sigma}}_{n,l}}{\gamma_{n,l}} \right)^{-1}  \hat{\vec{\mu}}_{n,l}\biggr\}. 
\end{aligned}
\label{equ:proof:mean:deri}
\end{equation}
Using the definitions (\ref{equ:prod:var}) and (\ref{equ:prod:mean}), we have
\begin{equation}
\begin{aligned}
&\frac{\partial J^{S}_{ini}(\vec{\mu}_w)}{\partial \vec{\mu}_w}=\sum_{n=1}^{N} 2\vec{\Theta}(\vec{s}_{n})
{\vec{\Sigma}_n^{S}}^{-1}
\vec{\Theta}^{\trsp}(\vec{s}_{n})\vec{\mu}_w \\
&-\sum_{n=1}^{N} 2\vec{\Theta}(\vec{s}_{n}){\vec{\Sigma}_{n}^{S}}^{-1} \vec{\mu}_{n}^{S}
=\frac{\partial \tilde{J}_{ini}^{S}(\vec{\mu}_w)}{\partial \vec{\mu}_w}. 
\end{aligned}
\label{equ:proof:mean:end}
\end{equation}
Thus, we conclude that the minimization problem in (\ref{equ:kl:mean:cost:mix}) is equivalent to the problem described by (\ref{equ:kl:mean:cost:mix:prod}).

\section{Proof of Weighted Variance Minimization Subproblem}
\label{app:compose:var}
The derivative of (\ref{equ:kl:var:cost:mix}) with respect to $\vec{\Sigma}_w$ is
\begin{equation}
\begin{aligned}
&\frac{\partial J^{S}_{ini}(\vec{\Sigma}_w)}{\partial \vec{\Sigma}_w}\!\!=\!\!\sum_{n=1}^{N}\!\sum_{l=1}^{L}\! \left\{-\gamma_{n,l}\vec{\Sigma}_w^{-1}\!+\!\gamma_{n,l}\vec{\Theta}(\vec{s}_n)\hat{\vec{\Sigma}}_{n,l}^{-1}\vec{\Theta}(\vec{s}_n)^{\trsp}\right\}\\
&=\!\!\sum_{n=1}^{N}\!\!\left\{\!\sum_{l=1}^{L}\!-\gamma_{n,l}\vec{\Sigma}_w^{-1}\!\right\} \!\!+\!\! \sum_{n=1}^{N}\! \vec{\Theta}(\vec{s}_{n})\!\biggl\{\!\sum_{l=1}^{L} \!\!
\left( \frac{\hat{\vec{\Sigma}}_{n,l}}{\gamma_{n,l}} \right)^{-1} \!\biggr\} \!\vec{\Theta}(\vec{s}_{n})^{\trsp}\\
&=\!\!\sum_{n=1}^{N}\{-\vec{\Sigma}_w^{-1}\}+\sum_{n=1}^{N} \vec{\Theta}(\vec{s}_{n}){\vec{\Sigma}_n^{S}}^{-1}  \vec{\Theta}(\vec{s}_{n})^{\trsp}\\
&=\frac{\partial \tilde{J}_{ini}^{S}(\vec{\Sigma}_w)}{\partial \vec{\Sigma}_w},
\end{aligned}
\label{equ:proof:var:end}
\end{equation}
so we have proved that the minimization problem in (\ref{equ:kl:var:cost:mix}) has the same solution as the problem defined in (\ref{equ:kl:var:cost:mix:prod}).
	
\end{appendices}	
	
\end{document}